\pdfoutput=1

\documentclass[11pt]{article}

\usepackage[final]{acl}

\usepackage{times}
\usepackage{latexsym}

\usepackage[T1]{fontenc}

\usepackage[utf8]{inputenc}

\usepackage{microtype}

\usepackage{inconsolata}

\usepackage{graphicx}

\title{Modeling Subjectivity in Cognitive Appraisal with Language Models}
\author{Yuxiang Zhou$^{1,2}$\thanks{Equal contribution},~~Hainiu Xu$^{2\ast}$,~~Desmond C. Ong$^{3}$,\\
\textbf{Maria Liakata}$^{1,4}$,~~\textbf{Petr Slovak}$^{2}$,~~\textbf{Yulan He}$^{2,4}$ \\
$^{1}$Queen Mary University of London,\quad$^{2}$King's College London\\
$^{3}$The University of Texas at Austin,~$^{4}$The Alan Turing Institute\\
\texttt{\{yuxiang.zhou,~m.liakata\}@qmul.ac.uk}\\
        \texttt{\{hainiu.xu,~petr.slovak,~yulan.he\}@kcl.ac.uk}\\
       	\texttt{desmond.ong@utexas.edu}\\
        }

\usepackage{graphicx}
\usepackage{amsmath,bm}
\usepackage{amsfonts}
\usepackage{array}
\usepackage{multirow}
\usepackage{enumitem}
\usepackage{booktabs}
\usepackage{microtype}
\usepackage{adjustbox}
\usepackage{tikz}
\usepackage{pgf}
\usepackage[dvipsnames]{xcolor}
\usepackage[most]{tcolorbox}
\definecolor{cornellred}{rgb}{0.7, 0.11, 0.11}

\newcommand{\argmin}{\mathop{\mathrm{argmin}}}

\definecolor{yellow}{HTML}{F6BD60}
\definecolor{white2}{HTML}{FFE0C1}
\definecolor{pink}{HTML}{F5CAC3}
\definecolor{tale}{HTML}{84A59D}
\definecolor{red}{HTML}{F28080}
\definecolor{green1}{HTML}{72C3A3}
\definecolor{green2}{HTML}{A5C2E2}
\definecolor{green3}{HTML}{70B48F}
\definecolor{orange}{HTML}{FE8019}
\definecolor{grey}{HTML}{EBDBB2}
\definecolor{brain}{HTML}{FFABBE}
\definecolor{blue}{HTML}{076678}
\definecolor{narrative}{HTML}{458588}

\newcolumntype{P}[1]{>{\centering\arraybackslash}p{#1}}

\newcommand{\tabbetter}[1]{\colorbox{green1}{#1}}
\newcommand{\tabworse}[1]{\colorbox{red}{#1}}
\newcommand{\tabsame}[1]{\colorbox{yellow}{#1}}

\newcommand{\improved}{\ensuremath{%
    \mathchoice{\includegraphics[height=1.4ex]{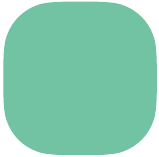}}
    {\includegraphics[height=1.4ex]{./figures/improved.pdf}}
    {\includegraphics[height=1.4ex]{./figures/improved.pdf}}
    {\includegraphics[height=1ex]{./figures/improved.pdf}}
}}

\newcommand{\unchanged}{\ensuremath{%
    \mathchoice{\includegraphics[height=1.4ex]{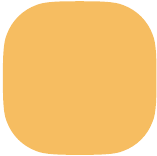}}
    {\includegraphics[height=1.4ex]{./figures/unchanged.pdf}}
    {\includegraphics[height=1.4ex]{./figures/unchanged.pdf}}
    {\includegraphics[height=1ex]{./figures/unchanged.pdf}}
}}

\newcommand{\worse}{\ensuremath{%
    \mathchoice{\includegraphics[height=1.4ex]{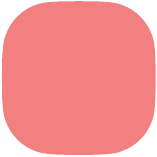}}
    {\includegraphics[height=1.4ex]{./figures/worse.pdf}}
    {\includegraphics[height=1.4ex]{./figures/worse.pdf}}
    {\includegraphics[height=1ex]{./figures/worse.pdf}}
}}

\begin{document}
\maketitle
\begin{abstract}
As the utilization of language models in interdisciplinary, human-centered studies grow, expectations of their capabilities continue to evolve.
Beyond excelling at conventional tasks, models are now expected to perform well on user-centric measurements involving confidence and human (dis)agreement- factors that reflect subjective preferences. 
While modeling subjectivity plays an essential role in cognitive science and has been extensively studied, its investigation at the intersection with NLP remains under-explored.
In light of this gap, we explore how language models can quantify subjectivity in~\textit{cognitive appraisal} by conducting comprehensive experiments and analyses with both fine-tuned models and prompt-based large language models (LLMs).
Our quantitative and qualitative results demonstrate that personality traits and demographic information are critical for measuring subjectivity, yet existing post-hoc calibration methods often fail to achieve satisfactory performance.
Furthermore, our in-depth analysis provides valuable insights to guide future research at the intersection of NLP and cognitive science\footnote{We make our code and resources available at \url{https://github.com/seacowx/CogApp-LLM-Subjectivity}.}.

\end{abstract}

\section{Introduction}

Large language models (LLMs) are increasingly deployed in high-stakes scenarios such as healthcare~\citep{sharma2023cognitive} and law~\citep{fan2024goldcoin} where the integration of human oversight is essential to mitigate potential risks. 
However, designing such systems poses significant challenges, as real-world human decision-making is prone to occasional errors and subjectivity~\cite{xiong2023can,zhou2023navigating,cheng2024measuring}.
Moreover, inherent disagreements among human annotators, stemming from the subjective nature of their judgments, further complicate the process~\citep{collins2023human,wang2024aligning}.
Most existing studies either assume the presence of a single human oracle, 
or aggregate multiple ratings using majority voting or averaging.
However, such methods fail to capture the nuances of human subjectivity.
Following~\citet{chen2008dimensions}, we define subjectivity as: \textit{the property that creates variances in experiences, interpretations, or behaviors shaped by internal states and personal perspective-encompasses multiple dimensions such as uncertainty, vagueness, and imprecision.} A focus on subjectivity aims to understand and model the inherent variability reflected in  individuals' internal states, which is fundamental to human reasoning about the world~\cite{o2006uncertain,lake2017building,chater2020probabilistic} and is indispensable for designing responsible AI systems~\citep{zhou2023navigating,cheng2024measuring}.

\begin{figure}[t!]
	\centering
	{\includegraphics[width=0.46\textwidth]{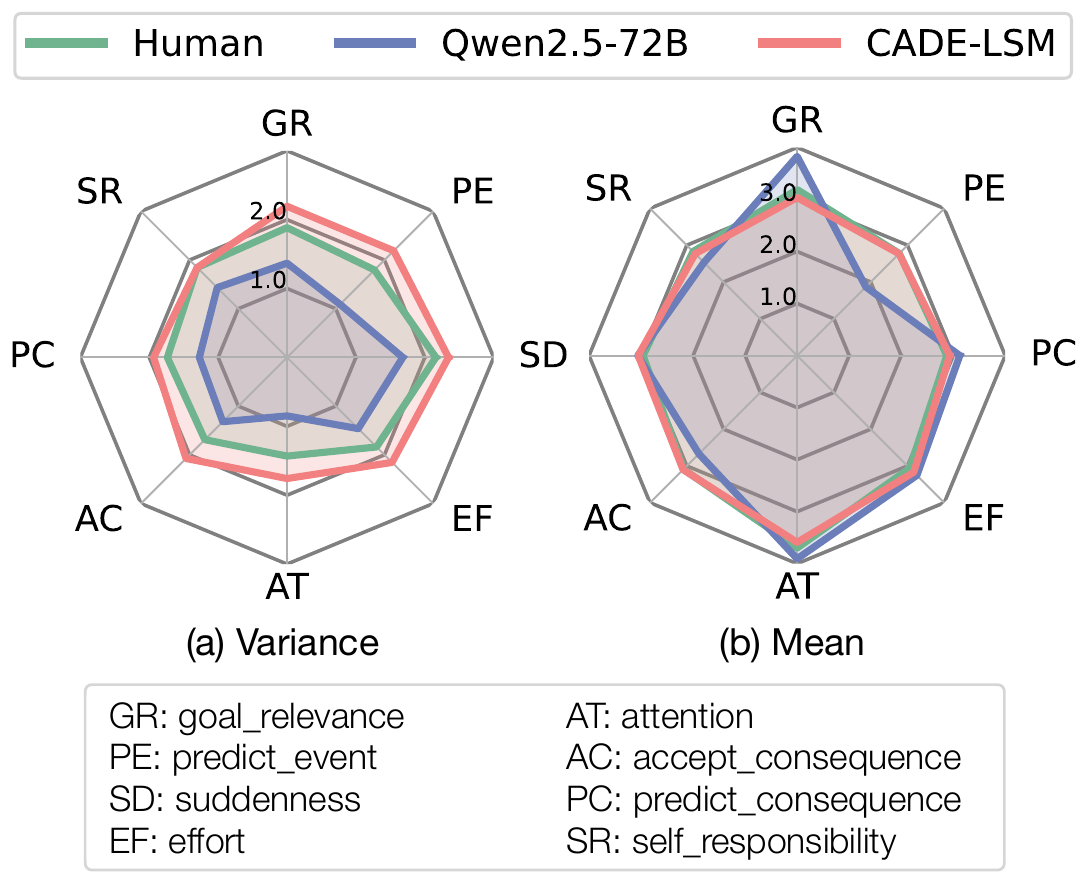}}
	\caption{Illustration of mean and variance of appraisal distribution between human and different models on 8 appraisal dimensions (Table~\ref{app:tab:appraisal_dimensions}). While models exhibit adequate capability in modeling the mean (b), they are lacking in modeling the variance (a).}
  \label{fig:intro}
\end{figure}

One important example of subjectivity is in how people experience and regulate emotions. 
Divergent emotional reactions may emerge from the same situation. In psychology, such variation is attributed to subjectivity in \emph{cognitive appraisal}, the process by which individuals evaluate and interpret events they experienced  in relation to their beliefs, goals, and prior experiences.~\citep{gross1998emerging,goldin2008neural,giuliani2009reappraisal,skerry2015neural,mcrae2016cognitive, yeo2024associations}. 
For example, when experiencing a romantic breakup, some individuals may blame themselves and appraise the situation to be "self-responsible" (causing emotions like \emph{guilt} or \emph{regret}) while others may appraise it to be "others-responsible" (causing emotions like \emph{anger}).
Recent work has examined the ability of LLMs to identify subjective appraisals in how individuals evaluate their situations~\citep{hofmann2020appraisal,zhan2023evaluating,yeo2023peace}, which in turn shapes their emotional experiences. 
Despite the progress achieved by these studies, 
existing methods often fail to capture the nuanced variability among individuals. 

As illustrated in Figure~\ref{fig:intro}, while models perform well in capturing average tendencies (mean), they struggle to model the inter-individual variance that is central to subjective interpretation. 
As a result, the models may end up representing only a narrow subset of individuals and lack the ability to properly model cognitive appraisal across varying population.
Modeling subjectivity enables user-centric evaluations that reflect individual preferences, which is crucial for the development of personalized and socially responsible systems.
For instance, LLMs can be guided to generate tailored re-appraisals for emotion, grounded in estimated appraisal variance~\cite{sharma2023cognitive,zhan2024large}.
While this concept has been extensively studied in cognitive science~\cite{schutz1942scheler,Demszky2023UsingLL,sharma2023human}, its investigation at the intersection with NLP remains under-explored~\citep{nlperspectives-2024-perspectivist}. 

To bridge this gap, we investigate how language models can be used to model subjectivity in cognitive appraisal.
Motivated by the increasing demand for human-centred evaluation and the growing interdisciplinary applications of LLMs, we frame our study around three key research questions: 1)~To what extent can language models quantify subjectivity in cognitive appraisal?
2)~Can their ability to measure subjectivity be improved, and if so, how? 3)~What insight can be gained from modeling subjectivity for practical applications?
To address these questions, we conducted a series of experiments and analyses across various scenarios using both fine-tuned models and prompt-based LLMs. In summary, our contributions are as follows:
\begin{itemize}[leftmargin=3mm]
    \item We conducted a pilot study to investigate how language models can be utilized to model subjectivity in cognitive appraisal.
    \item We explored methods to improve subjectivity quantification from two perspectives: knowledge injection and post-hoc calibration. Our findings suggest that personal profiles\footnote{\textbf{All personal profile information is anonymized and cannot be traced back to its provider.}}
    including personality traits and demographic information play a critical role in achieving better results, whereas current post-hoc calibration approaches often fail to produce satisfactory results.
    \item Our in-depth qualitative analysis provides valuable insights for future research at the intersection of NLP and cognitive science.
\end{itemize}

\section{Related Work}
\paragraph{Subjectivity Modeling.}

Existing NLP studies predominantly focus on modeling disagreement, aiming to reconcile conflicts among subjective viewpoints~\cite{pang2004sentimental, chen2008dimensions,uma2021learning,paun2021aggregating,plank2022problem,aher2023using}.
For example \citet{wang2024aligning} accounted for agreements among humans to train a preference model for natural language generation.
\citet{leonardelli2021agreeing} focused on the conflict of human annotators and investigated the impact of different degrees of disagreement.
However, subjectivity modeling seeks to leverage the inherent variability in individuals’ internal states and personal perspectives, thereby capturing personal and context-dependent nuances required for practical applications~\cite{shokri2024subjectivity,giorgi2024modeling}. 
The importance of subjectivity modeling stems not only from the growing deployment of language models in interdisciplinary research involving human-centred tasks and computational social science, but also from the need to ensure responsible and trustworthy human–AI interactions in high-stakes domains~\cite{gordon2022jury,giorgi2024modeling}. 
In light of this research gap, this paper explores foundational steps toward the development of language models capable of modeling subjectivity in cognitive appraisal.

\paragraph{Appraisal in Psychology.}
Understanding cognitive appraisal is essential for empowering individuals to regulate their emotions~\cite{gross1998emerging,goldin2008neural,giuliani2009reappraisal,skerry2015neural,mcrae2016cognitive}.
For instance, designing reappraisal interventions helps people change their interpretation of situations that trigger undesirable emotions~\cite{gross2015emotion}.
Psychological research has been focused on investigating the underlying cognitive mechanisms of this process.
\citet{uusberg2019reappraising} proposed \textit{reAppraisal} framework to explain these mechanisms in light of appraisal theory~\cite{scherer2001appraisal}.
Building on this work, \citet{uusberg2023appraisal} modeled the cognitive process involved in reappraisal aross various contexts by representing instances of reappraisal as profiles of shifts along abstract appraisal dimensions that characterize the significance of a situation for salient motives.
In a recent meta-analysis, \citet{yeo2024associations} identified a comprehensive list of 47 cognitive appraisal dimensions studied in across various theories.
In this paper, we followed~\citet{hofmann2020appraisal} to adapt 21 appraisal dimensions that have been widely examined in diverse scenarios in both psychological and NLP communities. 

\paragraph{Appraisal in NLP.}
Early efforts in appraisal analysis focused on classifying the appraisal dimensions conveyed in a given situation~\cite{hofmann2020appraisal}, for example, identifying the most likely appraisal from pre-defined dimensions, such as \textit{attention, coping, and responsibility}.
Another line of research focused on estimating the strength of people's appraisals in response to a situation they are experiencing.
For instance, \citet{zhan2023evaluating} and \citet{yeo2023peace} have introduced datasets to evaluate how LLMs assess cognitive appraisals. 
Other studies have explored how LLMs can assist people in reframing negative appraisals.
\citet{sharma2023cognitive} showed that highly specific and actionable appraisal reframing is considered the most helpful.
\citet{zhan2024large} suggested that LLMs can be guided to generate reappraisal response for emotional support with psychological principles.
Our work differs from previous studies in three key ways.
First, rather than focusing on a single appraisal dimension, we assume that multiple appraisal profile are underlying the appraisal process.
Second, instead of predicting point estimates, we model the underlying appraisal distribution, which allows us to capture subjectivity and variance.
Third, we evaluate our model in various real-world scenarios and find that people's appraisal pattern differ across situations.

\section{Modeling Subjectivity in Cognitive Appraisal Evaluation}

In psychology, Cognitive Appraisal Evaluation (CAE), the task of predicting human appraisal judgments on a given situation, is typically conducted by asking individuals to provide Likert-scale ratings\footnote{The human rater could be the event experiencer themselves or an observer of the event.} on a set of predefined appraisal dimensions related to an event or situation \cite{smith1985patterns}. 
Within the NLP community, researchers commonly frame CAE as a classification task, where human-provided appraisal ratings serve as ground truth labels \cite{zhan2023evaluating}.
However, numerous studies have shown that human appraisal ratings are not entirely consistent, even when assessing the same situation \cite{peacock1990stress, troiano2023dimensional}.
\begin{figure}[h]
	\centering
	\includegraphics[width=\columnwidth]{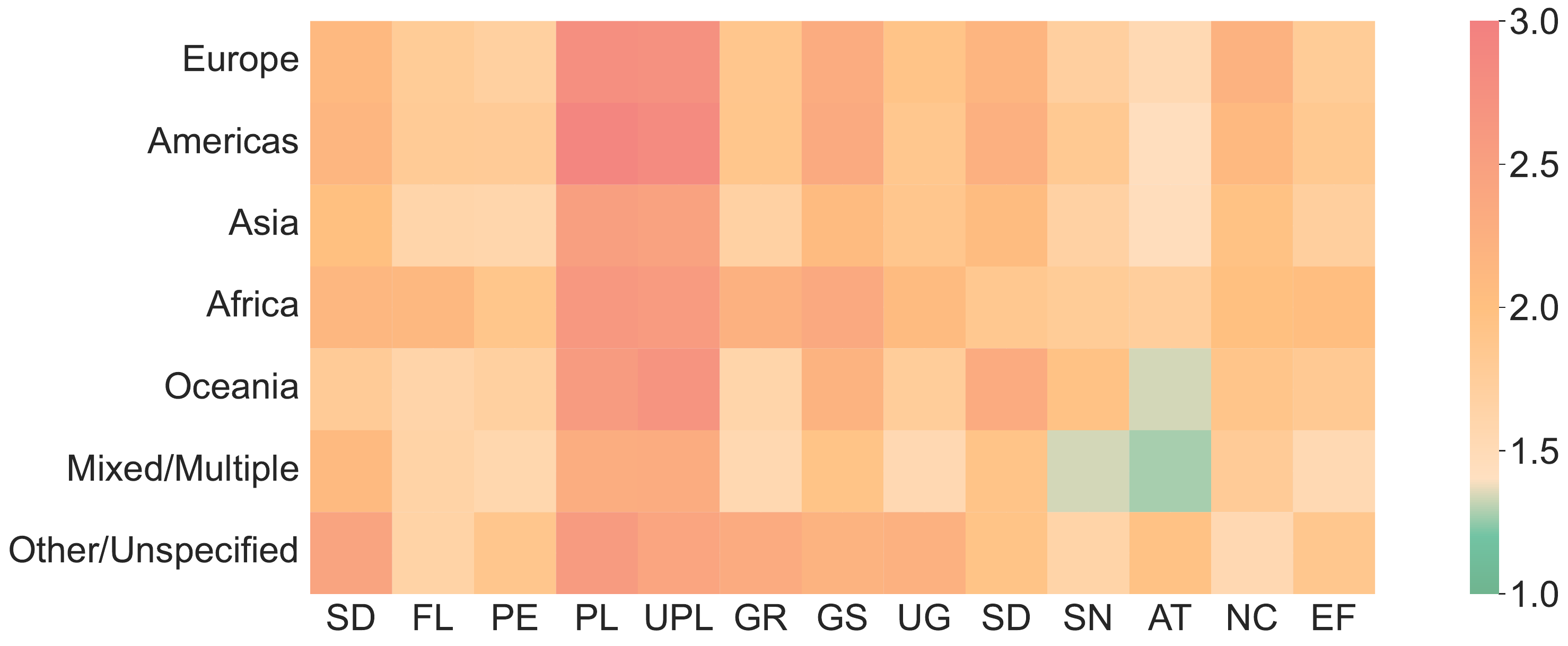}
	\caption{Appraisal variance across different geographic locations in the EnVent dataset \cite{hofmann2020appraisal}. The $x$-axis represents the appraisal dimensions as defined by \citet{hofmann2020appraisal}, while the $y$-axis represents the origins of the participants. See \S\ref{app:tab:appraisal_dimensions} for a detailed description of the appraisal dimensions.}
	\label{fig:appraisal_var}

\end{figure}

As illustrated in Figure~\ref{fig:appraisal_var}, individuals from different geographical regions  exhibit varying appraisal ratings for the same scenarios. From a micro perspective, variability in certain appraisal dimensions differs across geographical groups. For example, the variance of \textit{attention} (\texttt{AT}) among individuals from \textit{Oceania} is significantly lower than that of individuals from \textit{Africa}. From a macro perspective, the overall appraisal patterns also differ across geographical locations. For example, individuals from \textit{Mixed/Multiple} locations tend to have higher consensus across different appraisal dimensions compared to those from the \textit{Americas}.

To address the challenge of capturing the subjectivity inherent in human appraisal judgments, we adopt the Boltzmann policy for rating cognitive appraisals and propose a novel \textit{distribution-estimation} task for CAE. Formally, given a situation, $s$, and an appraisal dimension, $\delta_i \in \Delta$, the goal is to model the underlying distribution of the rating, $R_{\delta_i}$, $\mathbb{P}(R_{\delta_i} = r_{\delta_i} | s)$. Assuming that the rating at $\delta_i$ follows some latent distribution $d \in \mathcal{D}$, parameterized by $\theta \in \Theta$ (i.e. $R_{\delta_i} \sim d(\theta)$), the distribution estimation task can be formulated as:
\begin{equation}
	\hat{d}(\hat{\theta}) = \argmin_{d \in \mathcal{D}, \theta \in \Theta} \text{ dist}\left(\overline{d(\theta)}_{\delta_i}, d(\theta)\right),
\end{equation}
where $\text{dist}(\cdot, \cdot)$ is some distance metric between two probability distributions, and $\overline{d(\theta)}_{\delta_i}$ is the sample distribution obtained from repeated human measurements. 
Minding that different $R_{\delta_i}$
may follow distinct distributions, we do not make any assumption on the distribution $d(\theta)$. Instead of parameter estimation, an unorthodox task for language models, we construct sample distributions using language models and directly evaluate their quality.

We utilize two sets of metrics to evaluate the estimated distribution, namely a set of  point estimate metrics and a distribution metric.

\paragraph{Point Estimate Metrics} 
To evaluate the quality of the modeled distribution, a convenient measure is to assess the estimated mean and variance. We use sample mean and sample variance obtained from repeated measurements as the ground truth and compute the Mean Absolute Error (MAE) for both the estimated mean, $\mu$-MAE, and the estimated variance, $\sigma^2$-MAE.

\paragraph{Distribution Metric}
While the point estimates provide assessments of the quality of the estimated mean or variance, they do not reflect the holistic quality of the estimated distribution. 
To comprehensively evaluate the discrepancy between the sample distribution and the modeled distribution, a common measure is the KL divergence. However, KL divergence suffers from asymmetry and vulnerability to regions with zero probability density. As such, we opt to use the Wasserstein distance, which is a proper metric that measures the distance between two probability distributions using optimal transport plan \cite{kantorovich1960mathematical}. Formally, given two discrete probability distributions $p$ and $q$, the Wasserstein distance\footnote{We use the Wasserstein-1 distance.} is defined as
\begin{equation}
	W_1(p, q) = \int_{\mathbb{R}} |F_p(x) - F_q(x)| dx
\end{equation}
where $F_p(x)$ and $F_q(x)$ are the cumulative distribution function of $p$ and $q$ respectively. We report the average Wasserstein distance across all appraisal dimensions as the final evaluation score.

\section{Preliminary Explorations}

\label{sec:method}
To investigate language models' capability in modeling the subjectivity in CAE, we examine two categories of methods: fine-tuning pre-trained autoencoding language models (PLMs) and zero-shot prompting of autoregressive LLMs. 

\subsection{Fine-tuning of Pre-trained Models}

To tackle the distribution modeling task, we leverage the base model in two approaches, namely \textit{label-smoothing} and \textit{variational inference}.  

\paragraph{Label Smoothing} Originally introduced by \citet{szegedy2016rethinking} as a regularization method, \textit{Label Smoothing} has since been widely adopted in classification tasks. It reduces the kurtosis of the output logits by disseminating a portion of the probability density from ground truth label.
In distribution modeling, label densities ought to be allocated so that they resemble the the sample distribution. However, such a smoothing approach is infeasible as obtaining a sample distribution would require a large number of repeated measurements. 
Recognizing that 64\% of the appraisal ratings follow a unimodal distribution and 35\% follow a bimodal distribution\footnote{See \S\ref{app:human_rating_analysis} for detailed analysis of modalities of appraisal rating distributions}, we examine two label smoothing approaches. Firstly, we apply label smoothing using a discretized Gaussian distribution centered at the ground truth rating, which effectively reflect the unimodal distribution of appraisal ratings. Secondly, we  examine label smoothing with a mixture of two discretized Gaussian distributions, which replicate the bimodal distribution\footnote{Due to the subpar performance of the bimodal label smoothing approach, we defer the details to \S\ref{app:unimodal_vs_bimodal}.}.
We fine-tune PLMs using the smoothed labels as the target distribution. We refer to this approach as \textbf{CADE-LSM}. 

\paragraph{Variational Inference} The task of distribution-estimation has been widely studied in the machine learning community through variational inference \cite{jordan1999introduction, wainwright2008graphical}. From a Bayesian perspective, estimating the distribution in CAE is equivalent to inferring the posterior distribution of appraisal ratings given a situation. To this end, we adopt a Variational Autoencoder (VAE) \cite{Kingma2013AutoEncodingVB} to model the posterior distribution. Specifically, we use a PLM as the encoder to estimate the parameters of the latent distribution. Samples from the reparameterized latent distribution are then decoded using a two-layer MLP network to compute the evidence lower bound (ELBO). We additionally compute the Mean Squared Error (MSE) loss between the predicted and the ground truth appraisal ratings. 
The model is trained using the sum of the ELBO and the MSE loss. We refer to this approach as \textbf{CADE-VAE}.
Implementation details are in \S\ref{sec:apdx_cadevae}.

\subsection{Zero-shot Prompting}

Advancements in scaling autoregressive LLMs and aligning them with human preference data have enabled LLMs to mimic human communication and reasoning. As such, we explore whether LLMs can be used to model the subjectivity in CAE via zero-shot prompting. We investigate this under two settings. In the first setting, LLMs are tasked to provide an appraisal rating given a description of the situation and a description of a specific appraisal dimension\footnote{We follow~\cite{hofmann2020appraisal} to formulate the question for each appraisal dimension.}. For each situation, we prompt LLMs 30 times to obtain a sampled distribution of appraisal ratings. In the second setting, we additionally provide the LLMs with a personalized description, which include the Big-Five personality traits as well as demographic information. In this case, we prompt LLMs with the same situation paired with different personal profile to obtain a sampled distribution of appraisal ratings. Details of the prompts are provided in \S\ref{app:prompts}.

\paragraph{Finding the Optimal Temperature} 
When sampling appraisal ratings from LLMs, a key factor is the "\texttt{temperature}" parameter, which controls the randomness of the token generating process. To find an optimal temperature, we conduct a grid search over the temperature range $[0, 1.5]$. We select the temperature that yields the sampled distribution that is most similar to the distribution obtained from human ratings (\S\ref{app:tab:temperature_study}). 

\paragraph{Post-hoc Calibration} 
There has been bountiful studies that look into post-hoc methods for calibrating LLMs' confidence in their predictions \cite{linteaching, tian2023just}. While these methods emphasize the calibration of confidence (probability density) on the correct answer instead of the distribution over the label space, we examine whether these methods can be adopted to calibrate the distribution over all possible appraisal ratings. Specifically, we follow \citet{xiong2023can} and test two calibration methods: \textit{Average Confidence} and \textit{Pair Ranking} (\S\ref{sec:post_hoc_calib}). 

\section{Experiments}

To address our research questions, we systematically evaluate two categories of methods across three datasets: \underline{EnVent dataset}~\citep{hofmann2020appraisal} consists of daily event descriptions produced by native English speakers with 21 annotated appraisal dimensions.
\underline{FGE dataset}~\citep{skerry2015neural} includes 
descriptions of emotion-eliciting events with annotations along 38 appraisal dimensions.
\underline{CovidET dataset}~\citep{zhan2023evaluating} consists of situations described in Reddit posts related to COVID-19 and annotated along 24 appraisal dimensions.
Detailed statistics of the datasets can be found in Table~\ref{tab:stat}, and the full list of appraisal dimensions can be found in \S\ref{app:app_def}.
Further details on annotation validity can be found in \S\ref{app:quality_control}.

\begin{table}[htbp]
  \centering
      \resizebox{0.95\linewidth}{!}{
  \setlength{\tabcolsep}{3mm}{
    \begin{tabular}{lrccc}
    \toprule
    \textbf{Dataset} & Size  & Avg Len  & \# App Dim  & \# Ann \\
    \midrule
    EnVent & 1,200 & 111.2 & 21    & 5 \\
    FGE   & 200   & 291.1 & 14    & 8 \\
    CovidET & 40    & 727.3 & 16    & 2 \\
    \bottomrule
    \end{tabular}}}%
     \caption{Statistics of the evaluation datasets: \textit{Size}: dataset size / number of events; \textit{Avg Len}: average length of situation description in words; \textit{App Dim}: the number of appraisal dimensions; \textit{Ann}:  the number of human annotators for each event.}
  \label{tab:stat}%
\end{table}%

\begin{table*}[ht!]
	\centering
		\resizebox{\linewidth}{!}{
	\setlength{\tabcolsep}{2mm}{
	  \begin{tabular}{lccccccccc}
	  \toprule
	  \multicolumn{1}{c}{\multirow{2}[4]{*}{\textbf{Models}}} & \multicolumn{3}{c}{\textbf{EnVent}} & \multicolumn{3}{c}{\textbf{FGE}} & \multicolumn{3}{c}{\textbf{CovidET}} \\
	\cmidrule(lr){2-4}    \cmidrule(lr){5-7} \cmidrule(lr){8-10}       & Wasserstein~$\downarrow$  & $\mu$-MAE~$\downarrow$    & $\sigma^2$-MAE~$\downarrow$   & Wasserstein~$\downarrow$  & $\mu$-MAE~$\downarrow$    & $\sigma^2$-MAE~$\downarrow$   & Wasserstein~$\downarrow$  & $\mu$-MAE~$\downarrow$    & $\sigma^2$-MAE~$\downarrow$  \\
	\midrule
	Random & 1.196 & 1.096 & 1.060 & 1.191 & 1.088 & 1.042 & 1.438 & 1.367 & 0.833 \\
	Majority & 1.392 & 1.275 & 0.883 &  1.313 & 1.222 & 0.775 & 0.950 & 0.918 & 0.332 \\
    \midrule
	CADE-VAE & 1.279 & 0.984 & 0.882  & 1.209   & 1.105  & 0.713 & 1.200  & 1.106  & 0.331 \\
	CADE-LSM & \textbf{0.773} & \textbf{0.665} & 0.837 & \textbf{0.926} & \textbf{0.835} & 0.795 & 1.112 & 1.023 & 0.642 \\
        \midrule
	Llama3.1-8B & 1.094 & 0.904 & 0.826 & 1.409 & 1.385 & 0.780  & 1.109 & 1.065 & 0.361 \\
	Qwen2.5-7B & 1.078 & 0.919 & 0.817 & 1.131 & 1.020 & 0.704 & \textbf{0.905} & \textbf{0.864} & \textbf{0.322}  \\
  Llama3.3-70B & 1.012 & 0.926 & 0.820 & 1.248 & 1.203 & 0.740 & 0.970 & 0.924 & 0.325 \\
  Qwen2.5-72B & 0.945 & 0.852 & \textbf{0.736} & 1.048 & 0.960 & \textbf{0.632} & 0.905 & 0.867 & 0.325 \\
	\bottomrule
	\end{tabular}}}
	\caption{Main Results for the estimation of appraisal rating distribution. For all metrics, lower is better.}
	\label{tab:main}%

  \end{table*}%
  
\paragraph{Setup Details}
For fine-tuned models, we use DeBERTa-V3-Large as the backbone~\citep{he2021debertav3}.
Each model is trained for 30 epochs with a linear warmup for the first 10\% of the training steps. 
We employ AdamW~\cite{Loshchilov2017DecoupledWD} as the optimizer.
We set the maximum learning rate at 5e-5 with a batch size of 32 and select the optimal model weights based on MSE loss on the development set.
For in-domain experiments, we train on 4,680 examples from the EnVent dataset and select the best checkpoint using 540 validation examples.
For out-of-domain experiments, we fine-tune on EnVent training set and evaluate on the FGE and CovidET datasets.
Since the appraisal dimensions used in out-of-domain datasets slightly differ from those in EnVent, we manually aligned their dimensions and ratings for fair comparision.
Further details are provided in \S\ref{app:app_def}.

\paragraph{Baselines}
The baseline models selected for comparison can be broadly categorized into two groups: fine-tuned models: \textbf{CADE-LSM} estimates subjectivity using label-smoothing~\citep{rolf2022resolving,wang2024aligning}; \textbf{CADE-VAE}, a latent variable model infers appraisal distribution using VAE~\citep{Kingma2013AutoEncodingVB}.
Prompt-based models include:~\textbf{Llama3.1-8B} and \textbf{Llama3.3-70B}~\cite{dubey2024llama};  \textbf{Qwen2.5-7B} and \textbf{Qwen2.5-72B}~\cite{yang2024qwen2}.

\subsection{Results and Analysis}
\label{sec:results_and_analysis}
\subsubsection*{How well do language models quantify subjectivity in cognitive appraisal ratings?}
\paragraph{\textit{Fine-tuning PLMs brings significant improvement in mean estimation.}} 
As discussed in \S\ref{sec:method}, we use the Wasserstein distance as the primary metric for its comprehensiveness. Point estimate metrics are utilized to provide insights into two aspects of the estimated distribution: mean and variance.

As shown in Table~\ref{tab:main}, fine-tuning on the EnVent dataset with label smoothing consistently outperforms all baseline models in terms of Wasserstein distance on both EnVent and FGE dataset.
For instance, CADE-LSM surpasses the second-best baseline, Qwen2.5-7B, by 39.5\% in Wasserstein distance. However, when examining the point estimate of the mean and variance, we see that the improvement in Wasserstein distance is largely attributed to the improvement in mean estimation. In terms of variance estimation, CADE-LSM underperforms all LLMs, suggesting that while fine-tuning PLMs can effectively capture the \textit{general tendency} of human appraisal ratings (as reflected in the mean), it struggles to model the subjectivity in appraisal ratings (as reflected in the variance).

Moreover, fine-tuned models suffer from limited generalizability.
For instance, trained with EnVent, in which the situations are described with concise sentences, CADE-LSM generalizes well to the FGE datasets, which contains situation descriptions of similar kind\footnote{EnVent and FGE datasets contain daily event descriptions.}, but underperforms the Majority baseline on the CovidET dataset, which contains reddit posts that are longer in length and focus on COVID-19 related events.

\paragraph{\textit{LLMs are more effective in modeling variance in appraisal ratings.}}
Although LLMs are less effective compared to fine-tuned PLMs in terms of the Wasserstein distance, they are more adept at modeling the variance in appraisal ratings. As shown in Table~\ref{tab:main}, the Qwen2.5 family consistently achieves the best point estimate for variance ($\sigma^2$-MAE). Similar to the variance-bias tradeoff, LLMs are good at modeling the variance while failing at reducing the bias. However, such results are insufficient in proving LLMs' effectiveness in modeling variance since the randomness of LLM generations can be easily adjusted by altering the temperature parameter (Table~\ref{app:tab:temperature_study}). To obtain a concrete conclusion, results in the following controlled experiments show that incorporating personal profile in the prompt significantly improves the estimated variance, especially for the Qwen2.5 family of models (Table~\ref{tab:rq2_persona_abl}). 
We also observe differences in behavior with respect to model scale.
In general, large models achieved stronger overall performance.
However, these improvements were not uniformly distributed, and no single model consistently outperformed others across all settings and datasets.

\subsubsection*{Does personal profile help language models capture subjectivity?}
Studies in psychology have shown that personal profiles such as personality traits are critical to shaping human appraisal judgement \cite{mischel1995cognitive, childs2014personality}. To understand the role of personal profile in modeling subjectivity with language models, we incorporate two types of information: demographic information and the Big-Five personality traits~\cite{gosling2003very}. 
\paragraph{\textit{Effectiveness of personal profile varies across models.}}

As shown in Table~\ref{tab:rq2_persona_abl}, while models incorporated with personal profile generally achieve better performance in terms of Wasserstein distance, the degree of performance increment is model-dependent.
The introduction of personal profiles generally leads to better performance for lLMs compared to fine-tuned PLMs, although the latter still achieve marginal improvements in certain scenarios.
Specifically, we observe that adding personal profile is particularly effective for Qwen2.5-7B, which shows improvement across all metrics. 
In contrast, Llama3-8B only shows slight improvements in Wasserstein distance and $\sigma^2$-MAE, but decreased performance in $\mu$-MAE.

\begin{table}[t!]
  \centering
        \resizebox{\linewidth}{!}{
  \setlength{\tabcolsep}{2mm}{
    \begin{tabular}{lccc}
    \toprule
    \multicolumn{1}{c}{\textbf{Models}} & Wasserstein~$\downarrow$ & $\mu$-MAE~$\downarrow$   & $\sigma^2$-MAE~$\downarrow$ \\
    \midrule
    CADE-VAE & 1.279 & 0.984 & 0.882 \\
    \textcolor{white}{aa}~w.~Demo & \tabsame{1.281} & \tabsame{0.983} & \tabsame{0.882} \\
    \textcolor{white}{aa}~w.~Traits & \tabsame{1.281} & \tabsame{0.983} & \tabsame{0.882} \\
    \textcolor{white}{aa}~w.~Demo \& Traits & \tabsame{1.279} & \tabsame{0.984} & \tabsame{0.882} \\
    \midrule
    CADE-LSM & 0.773 & 0.665 & 0.837 \\
    \textcolor{white}{aa}~w.~Demo & \tabbetter{0.768} & \tabworse{0.682} & \tabworse{0.843} \\
    \textcolor{white}{aa}~w.~Traits & \tabbetter{0.763} & \tabworse{0.676} & \tabbetter{0.832} \\
    \textcolor{white}{aa}~w.~Demo \& Traits & \tabbetter{0.765} & \tabworse{0.678} & \tabbetter{0.818} \\
    \midrule
    \midrule
    Llama3-8B & 1.094 & 0.904 & 0.826 \\
    \textcolor{white}{aa}~w.~Demo & \tabsame{1.090} & \tabworse{1.014} & \tabsame{0.823} \\
    \textcolor{white}{aa}~w.~Traits & \tabworse{1.148} & \tabworse{1.078} & \tabworse{0.848} \\
    \textcolor{white}{aa}~w.~Demo \& Traits & \tabbetter{1.086} & \tabworse{1.008} & \tabbetter{0.817} \\
    \midrule
    Qwen2.5-7B & 1.078 & 0.919 & 0.817 \\
    \textcolor{white}{aa}~w.~Demo & \tabbetter{0.994} & \tabbetter{0.845} & \tabbetter{0.799} \\
    \textcolor{white}{aa}~w.~Traits & \tabbetter{0.962} & \tabbetter{0.821} & \tabbetter{0.775} \\
    \textcolor{white}{aa}~w.~Demo \& Traits & \tabbetter{0.958} & \tabbetter{0.826} & \tabbetter{0.756} \\
    \midrule
    Llama3-70B & 1.012 & 0.926  & 0.820 \\
    \textcolor{white}{aa}~w.~Demo & \tabbetter{0.992} & \tabbetter{0.872} & \tabworse{0.828} \\
    \textcolor{white}{aa}~w.~Traits & \tabbetter{0.979} & \tabbetter{0.871} & \tabbetter{0.805} \\
    \textcolor{white}{aa}~w.~Demo \& Traits & \tabbetter{0.967}  & \tabbetter{0.841}  & \tabbetter{0.811} \\
    \midrule
    Qwen2.5-72B & 0.945 & 0.852  & 0.736 \\
    \textcolor{white}{aa}~w.~Demo & \tabworse{0.961} & \tabbetter{0.838} & \tabsame{0.738} \\
    \textcolor{white}{aa}~w.~Traits & \tabsame{0.946} & \tabbetter{0.828} & \tabbetter{0.724} \\
    \textcolor{white}{aa}~w.~Demo \& Traits & \tabsame{0.948}  & \tabbetter{0.836}  & \tabbetter{0.714} \\
    \bottomrule
    \end{tabular}}}%
      \caption{Personal profile study with various models on EnVent dataset. $\improved$ Improved, $\unchanged$ Unchanged, and $\worse$ Decreased results are highlighted.}
  \label{tab:rq2_persona_abl}%

\end{table}%

\paragraph{\textit{Different types of personal profile contribute differently to model performance.}}

To understand the contribution of each type of personal profile, we examine all models with either demographic (e.g., gender) information or personality traits (e.g., extraversion).
As shown in Table~\ref{tab:rq2_persona_abl}, we found that adding personal profile information brings contrasting effects on fine-tuned models: while CADE-VAE does not benefit from the additional information, CADE-LSM is able to better capture the subjectivity in the appraisal ratings, which is evident by its improved \textit{Wasserstein} and $\sigma^2$-MAE. 
For prompt-based models, results are mixed. Incorporating personality traits or demographic information improves mean modeling ($\mu$-MAE) for all LLMs except Llama3-8B, but this does not consistently yield better \textit{Wasserstein} scores, as observed with Qwen2.5-72B.

\paragraph{\textit{Significant improvement from personal profiles integration.}} To evaluate the impact of incorporating personal profiles on the subjectivity of various appraisal dimensions, we conducted a one-tailed two-sample T-test.
Comprehensive p-value results are shown in Table~\ref{tab:sig_profile}. 
Our results show that 1/3 of the appraisal dimensions exhibit statistically significant improvements when either personality traits or demographical information are integrated, across both Llama3.1-8B and Qwen2.5-7B.
For example, when both types of personal profiles are integrated, significant gains are observed for \textit{predict\_event, self\_control, accept\_consequence}, and \textit{effort}.
Detailed analyzes can be found in \S\ref{sec:apdx_sg}.
These findings reassure that incorporating personal profiles play a vital role in modeling subjectivity.

\subsubsection*{Do existing post-hoc calibration methods improve the modeling of subjectivity?}
\label{sec:post_hoc_calib}

\paragraph{\textit{Post-hoc calibration degrades model performance.}} We conducted experiments using Avg-Conf and Pair-Rank methods \cite{xiong2023can}. In the setting of CAE, Avg-Conf samples multiple (rating confidence) pairs from LLMs by explicitly instructing LLMs to output a confidence score associated with each rating. Calibration is done by normalizing the confidence scores across all ratings. Pair-Rank assumes that LLMs are better at ranking a given set of ratings. By sampling multiple rankings of the ratings, pair-rank computes a categorical distribution over the rating space using stochastic gradient descent to optimize a conditional log-likelihood function based on the sampled rankings. 

\begin{table}[h]
  \centering
        \resizebox{\linewidth}{!}{
  \setlength{\tabcolsep}{2mm}{
    \begin{tabular}{lccc}
    \toprule
    \multicolumn{1}{c}{\textbf{Models}} & Wasserstein~$\downarrow$ & $\mu$-MAE~$\downarrow$   & $\sigma^2$-MAE~$\downarrow$ \\
     \midrule
     Llama3-8B & 1.094 & 0.904 & 0.826 \\
    \textcolor{white}{Llama}~w. Avg-Conf & \tabworse{1.127}  & \tabworse{1.022}  & \tabbetter{0.804} \\
    \textcolor{white}{Llama}~w. Pair-Rank  & \tabworse{1.768}  & \tabworse{1.752}  & \tabbetter{0.788} \\
    \midrule
    \midrule
    Qwen2.5-7B & 1.078 & 0.919 & 0.817 \\
    \textcolor{white}{Qwen}~w. Avg-Conf & \tabbetter{1.045}  & \tabbetter{0.873} & \tabworse{0.828} \\
    \textcolor{white}{Qwen}~w. Pair-Rank & \tabworse{1.167} & \tabworse{1.081} & \tabbetter{0.797} \\
    \bottomrule
    \end{tabular}}}%
    \caption{Post-hoc calibration on EnVent dataset. Lower values are better. $\improved$ Improved and $\worse$ Decreased results are highlighted in corresponding color.}
  \label{tab:rq2_model_abl}%
\end{table}%

Results shown in Table~\ref{tab:rq2_model_abl} demonstrate that the Avg-Conf and Pair-Rank calibration strategies lead to limited improvement in the case of Qwen2.5-7B and degraded performance in Llama3-8B.
One possible explanation is that existing calibration methods are designed for classification setup, where the objective is to align the logit of a predicted class with the probability of that the class being correct \cite{guo2017calibration}.
This setup is fundamentally different from the distribution estimation task in CAE, where the probability density allocated to each rating  must be modeled, which is much more intricate than the classification setup.
Further research is needed to investigation the underlying reasons. 

\subsubsection*{Qualitative Study: Understanding Subjectivity}%

Table~\ref{tab:quali_ad} presents a qualitative analysis of subjectivity measurements for different models across various appraisal dimensions in the EnVent dataset. We observe slight differences in the top-three and bottom-three dimensions between fine-tuned and prompt-based models. 
For instance, CADE-LSM aligns most closely with human subjectivity on the dimensions of~\textit{pleasantness} and least on \textit{urgency}, whereas Lama3.1-8B shows the strongest alignment on \textit{social\_norms} and the weakest on \textit{other\_control}. 
Nonetheless, we found that the dimensions of \textit{pleasantness}, \textit{social\_norms}, and \textit{unpleasantness} are captured most effectively by these models, while \textit{familiarity}, \textit{goal\_support}, and \textit{other\_control} remain challenging for all models. 
Interestingly, we found that personal profile does not affect the qualitative results of subjectivity across all models.
Comprehensive experiments are provided in \S\ref{sec:apdx_qs}.
These findings highlight the need for future research to accurately quantify various dimensions of subjectivity.

\begin{table}[t]
  \centering
   \resizebox{\linewidth}{!}{
  \setlength{\tabcolsep}{3mm}{
    \begin{tabular}{lccc}
    \toprule
    \multirow{2}[4]{*}{\textbf{Models}} & \multicolumn{3}{c}{Top Quantified Appraisal Dimensions} \\
\cmidrule{2-4}          & Rank 1 & Rank 2 & Rank 3 \\
    \midrule
    CADE-LSM & \textit{pleasantness} & \textit{unpleasantness} & \textit{social\_norms} \\
    Lama3-8B & \textit{social\_norms} & \textit{pleasantness} & \textit{unpleasantness} \\
    Qwen2.5-7B & \textit{pleasantness} & \textit{social\_norms} & \textit{unpleasantness} \\
    Llama3.3-70B & \textit{pleasantness} & \textit{unpleasantness} & \textit{social\_norms} \\
    Qwen2.5-72B & \textit{social\_norms} & \textit{suddenness} & \textit{effort} \\
    \midrule
    \midrule
    & \multicolumn{3}{c}{Bottom Quantified Appraisal Dimensions} \\
    \midrule
    CADE-LSM & \textit{urgency} & \textit{goal\_support} & \textit{predict\_consequence} \\
    Llama3.1-8B & \textit{other\_control} & \textit{goal\_support} & \textit{familiarity} \\
    Qwen2.5-7B & \textit{familiarity} & \textit{goal\_support} & \textit{other\_control} \\
    Llama3.3-70B & \textit{other\_control} & \textit{attention} & \textit{accept\_consequence} \\
    Qwen2.5-72B & \textit{familiarity} & \textit{urgency} & \textit{self\_responsibility} \\
    \bottomrule
    \end{tabular}}}%
      \caption{Qualitative analysis for different models on various appraisal dimensions. Wasserstein distance is used as the measurement metric.}
  \label{tab:quali_ad}%
\end{table}%
\begin{figure}[b!]
	\centering
	\includegraphics[width=\columnwidth]{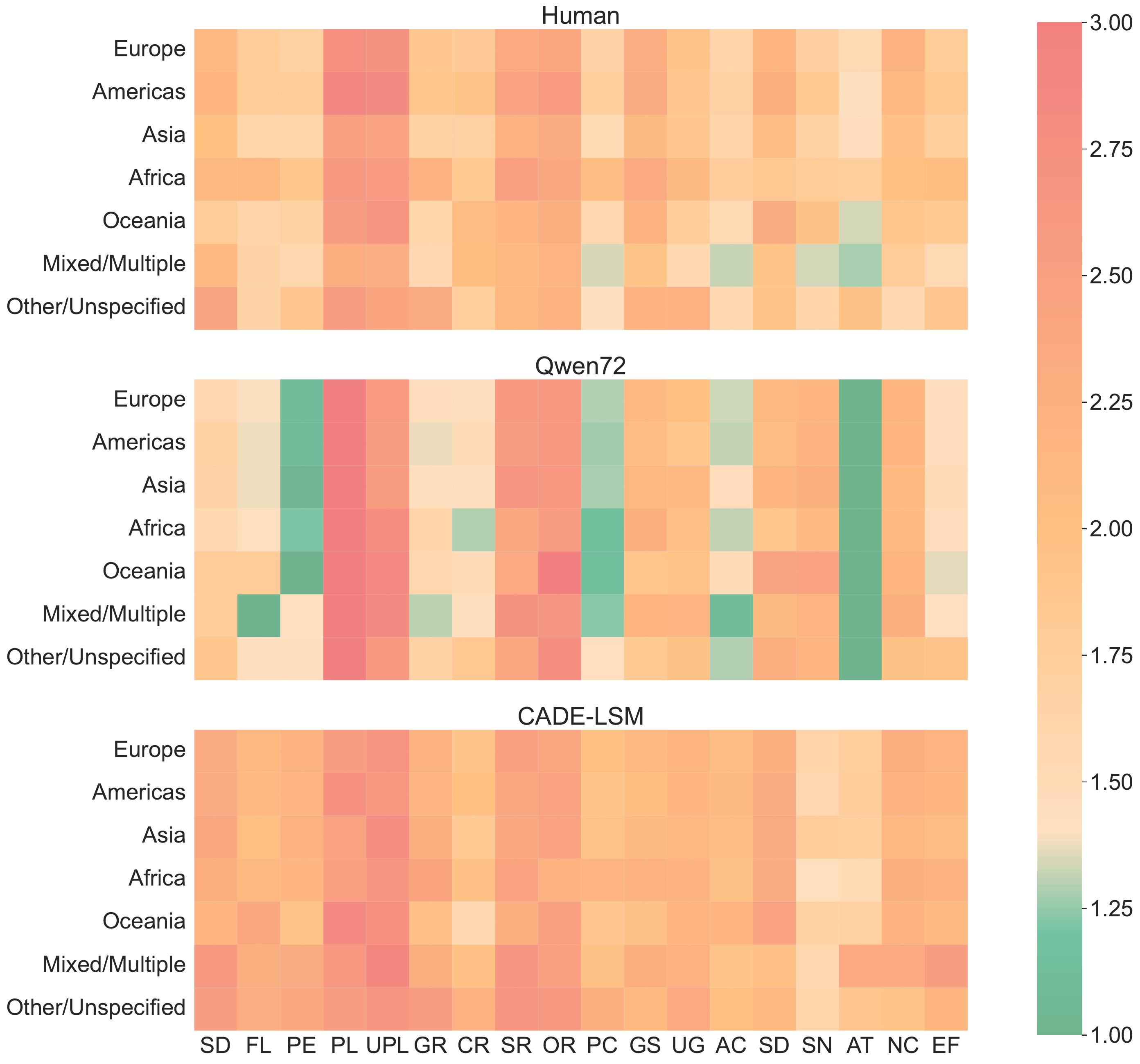}
	\caption{Comparison of appraisal variance in different geographic locations in the EnVent dataset. See Table~\ref{app:app_def} for definition of the abbreviated appraisal dimensions.}
	\label{fig:appraisal_var_comp}
\end{figure}
\paragraph{Effects of Demographics}
To gain further insight on the effects of different demographic on the subjectivity of cognitive appraisal, we inspect the variance in different geographic locations in the EnVent dataset in Figure \ref{fig:appraisal_var_comp}.
The $x$-axis represents the appraisal dimensions defined in \citet{hofmann2020appraisal}. The $y$-axis represents the origins of the participants.
As shown in Figure \ref{fig:appraisal_var_comp}, CADE-LSM exhibits subjectivity patterns that align more closely with human judgments across all appraisal dimensions, suggesting substantial disagreement among individuals of different geographic locations in each dimension. 
Specifically, individuals show considerable variability in dimensions such as \textit{pleasantness} (\texttt{PL}), \textit{unpleasantness} (\texttt{UPL}), \textit{self\_responsibility} (\texttt{SR}), and \textit{other\_responsibility} (\texttt{OR}).
In contrast, the Qwen2.5-72B model provides more consistent subjectivity quantification for individuals from different geographic locations in dimensions like \textit{predict\_event} (\texttt{PE}) and \textit{predict\_consequence} (\texttt{PC}).
We deduce that, compared to prompt-based LLMs, fine-tuning allows the model to learn contextualized personal profile,  thereby enabling them to more accurately capture subjectivity.

Upon further investigation, our study indicates that individuals with various geographic locations exhibit relatively consistent subjective cognition regarding \textit{social\_norms} (\texttt{SN}) and \textit{familiarity} (\texttt{FL}), yet responsibility-related appraisal dimensions (e.g., \texttt{CR}, \texttt{SR}, \texttt{OR}) provoke more pronounced variability.

\paragraph{Effects of Personality Traits}
We conduct a similar analysis focusing on personality traits.
We again find that fine-tuned models exhibit appraisal patterns closely align with those of human, indicating the critical role of personal profile in modeling subjectivity with fine-tuned models.
Figure~\ref{fig:trait_var} provides further details on the comparison.

In summary, we found that certain appraisal dimensions such as \textit{pleasantness} and \textit{unpleasantness} show inherent variance regardless of individuals’ geographic locations or personality traits.
In contrast, dimensions such as \textit{attention} and \textit{accept\_consequence} exhibit relatively small variations. 
We hypothesize that sentiment-related appraisal dimensions tend to elicit more extreme ratings, reflecting higher subjectivity than other dimensions.
This effect may be driven by the annotation strategy used in EnVent dataset~\citep{hofmann2020appraisal}, which naturally incorporates bi-polar sentiment and thus amplifies variance.
More discussion can be found in \S\ref{sec:apdx_qs}.
Further research is needed to investigate the underlying factors driving these findings and to develop more advanced techniques for improving models’ ability to model subjectivity in cognitive appraisal.

\section*{Practical Implications}
While mean-based measurements capture how representative a model is of a population, variance reflects how subjective a particular dimension may be within a population.
In our experiments, we found only week correlations between these two types of metrics, and the relationship was model-dependent.
This suggests that, beyond building models that representative of a population, it is equally important to develop models capable of capturing the nuanced subjectivity within each population.
More detailed analysis can be found in \S\ref{app:mean_var_corr}.
Building on our exploration of first two research question, we now offer practical recommendations for measuring subjectivity that are both psychologically meaningful and computationally feasible.
\subsection*{Computational perspective}
First, metric selection should be task- or dataset-dependent.
For example, datasets share similar domains and event contexts tend to yield similar ranges of $\sigma$-MAE scores, vice versa.
However, we argue that the Wasserstein distance provides a holistic measure for evaluating whether LLMs approximate human judgement distributions in subjectivity tasks, as it captures both mean and variance. 
In practice, a model that poorly estimates the mean will also fail to represent the distribution meaningfully.
Second, model choice depends on data availability. 
When sufficient annotated data is avaiable, we recommend fine-tuned models such as CADE-LSM, which perform better at holistic measurement of subjectivity.
In low-resource setting, where annotations are sparse or unavailable, prompt-based LLMs can serve as a reasonable proxy with better generalizability, specifically for dimensions that benefit from personal profile integraiton, such as \textit{predict\_event, self\_control}, and \textit{accept\_consequence}.
\subsection*{Psychological Perspective}
Certain appraisal dimensions are well modeled by either fine-tuned PLMs or prompt-based LLMs.
Dimensions such as \textit{pleasantness, unpleasantness}, and \textit{self\_responsibility} show high variance among annotators, yet LLMs capture their subjectivity reasonably well.
By contrast, \textit{goal\_support} exhibits high variance, but models struggle to replicate it.
In the same time, \textit{social\_norms} displays low variance in human ratings, and this stability is likewise well modeled by LLMs.
However, \textit{predict\_consequence} shows low human variance but is poorly captured by models, suggesting that low variance alone does not guarantee replicability.

\section{Conclusion}
In this paper, we take an important first step toward modeling subjectivity in cognitive appraisal with language models.
We conduct detailed experiments and analysis across various scenarios with both fine-tuned PLMs and prompt-based LLMs.
Our findings reveal notable inconsistencies in modeling subjectivity, with no single model reliably identifying it across all appraisal dimensions under varying conditions.
The thorough quantitative and qualitative examination indicates that personality traits and demographical information play a vital role in measuring subjectivity, whereas existing post-hoc calibration approaches fail to achieve improved results.
Furthermore, our qualitative analysis provides valuable insights for future research and development in the intersection of NLP, Cognitive Modeling, and Psycholinguistics.

\section{Limitation}
While this paper lays the groundwork for further exploration into subjectivity modeling, it is necessary to acknowledge its limitations.
First, we treated appraisal dimensions independently, without consideration of their potential interrelationships.
However, certain dimensions are closely correlated.
For example, \textit{self\_control}, \textit{other\_control}, and \textit{chance\_control} are all fall in the coping objective addressing control attributions~\cite{sander2005systems,scherer2013driving,troiano2023dimensional}.
By incorporating prior knowledge of correlations among these dimensions, the appraisal distribution could be more accurately modeled for subjectivity.
Secondly, the datasets used in our experiments involve a limited number of annotators, particularly the CovidET dataset, which included only two annotators per sample.
We observed that in datasets with a moderate number of annotators (e.g., FGE and EnVent, where $n\geq5$), model predictions exhibited similar trends, whereas the CovidET dataset showed different model behavior.
To enhance statistical robustness, future work should incorporate a larger pool of annotators (e.g., $n\geq30$).
Thirdly, we employed relatively simple prompts for LLMs, whereas prior research suggests that more advanced prompting methods could enhance model performance~\cite{wei2022chain,zhou2024mystery}.
Moreover, our experiments focused primarily on English, meaning the findings may not necessarily generalize to multilingual contexts~\cite{banea2008multilingual,banea2010multilingual}. Investigating subjectivity modeling across different languages remains an avenue for future research.
Finally, LLMs exhibit biases that can lead to unfaithful appraisal ratings, especially when persona profiles are involved~\citep{wang2024large,dong2024persona}. Further work is needed to better align LLMs with provided profiles to ensure faithful and unbiased appraisal ratings.

\section{Ethical Considerations}
The primary goal of this study is to draw attention of quantifying inherent subjectivity in cognitive appraisal in future studies.
All datasets used originate from previously published work that have undergone thorough ethical review to ensure safe and responsible use~\cite{hofmann2020appraisal,skerry2015neural,zhan2023evaluating}.
Any harmful or offensive content was screened and removed.
Also, there is no personally identifiable information presented in the data, and all named entities have been masked to ensure privacy.

In this work, several personality traits were reduced to binary for ease of analysis.
This leads our findings to a limited coverage of the population and introducing a degree of selection bias commonly observed in real-world scenarios.
We emphasise that we do not intend to constrain or imply these per by the limited definitions employed in this paper.
While the work presented here facilitated a relatively straightforward estimation of subjectivity, it also raises potential concerns regarding privacy when incorporating personal profile.
We believe that future application may benefit from integrating a theory-of-mind perspective~\citep{xu2025enigmatom}, thereby reducing reliance on explicit personal data.
\section*{Acknowledgements}
We would like to thank the anonymous reviewers for their constructive comments and suggestions.
We are grateful to Amy E. Skerry and Prof. Rebecca Saxe for sharing the FGE dataset, and to Ashish Mehta and Kate Petrova for their helpful discussions.
This work was supported in part by the UK Engineering and Physical Sciences Research Council (EPSRC) through a Turing AI Fellowship (grant no. EP/V020579/1, EP/V020579/2), as well as through UKRI Future Leaders fellowship (grant no. MR/Y034295/1 and MR/T041897/1) and the Engineering and Physical Sciences Research
Council [grant number EP/Y009800/1], through
funding from Responsible Ai UK (KP0016).

\bibliography{anthology.arxiv,custom}
%

\clearpage
\appendix
\setcounter{table}{0}
\renewcommand{\thetable}{A\arabic{table}}
\setcounter{figure}{0}
\renewcommand{\thefigure}{A\arabic{figure}}

\begin{table*}[ht]
	\centering
		\resizebox{0.95\linewidth}{!}{
	\setlength{\tabcolsep}{2mm}{
	  \begin{tabular}{l l l}
	  \toprule
        Appraisal Dimension & Abbrev. & Description \\
        \midrule 
        suddenness & SD & The event was sudden or abrupt. \\
        familiarity & FL & The event was familiar to its experiencer. \\
        predict\_event & PE & The experiencer could have predicted the occurrence of the event. \\
        pleasantness & PL & The event was pleasant for the experiencer. \\
        unpleasantness & UPL& The event was unpleasant for the experiencer. \\
        goal\_relevance & GR & The experiencer expected the event to have important consequences for him/herself. \\
        chance\_responsibility & CR & The event was caused by chance, special circumstances, or natural forces. \\
        self\_responsibility & SR & The event was caused by the experiencer’s own behavior. \\
        other\_responsibility & OR & The event was caused by somebody else’s behavior. \\
        predict\_consequence & PC & The experiencer anticipated the consequences of the event. \\
        goal\_support & GS & The experiencer expected positive consequences for her/himself. \\
        urgency	& UG & The event required an immediate response. \\
        self\_control & SC & The experiencer expected positive consequences for her/himself. \\
        other\_control & OC & Someone other than the experiencer was inﬂuencing what was going on. \\
        chance\_control & CC & The situation was the result of outside inﬂuences of which nobody had control. \\
        accept\_consequence & AC & The experiencer anticipated that he/she could live with the unavoidable consequences of the event. \\
        standards & SD & The event clashed with her/his standards and ideals. \\
        social\_norms & SN & The actions that produced the event violated laws or socially accepted norms. \\
        attention & AT & The experiencer had to pay attention to the situation. \\
        not\_consider & NC & The experiencer wanted to shut the situation out of her/his mind. \\
        effort & EF & The situation required her/him a great deal of energy to deal with it. \\
	\bottomrule
	\end{tabular}}}
	\caption{Definition of appraisal dimensions in \citep{hofmann2020appraisal}}
	\label{app:tab:appraisal_dimensions}%
\end{table*}

\section{Detail of Appraisal Dimensions}
\label{app:app_def}
\subsection{Definition of the 21 Appraisal Dimensions}
There are various definitions of cognitive appraisal dimensions. In this work, we use the 21 appraisal dimensions defined in \citep{hofmann2020appraisal}. The definition of the 21 appraisal dimensions as listed in Table~\ref{app:tab:appraisal_dimensions}.

\begin{table} [h!]
  \centering 
  \resizebox{\columnwidth}{!}{
      \begin{tabular}{P{0.01\columnwidth} l l}
          \toprule
          & Original & EnVent \\
          \midrule
          \multirow{14}{*}{\rotatebox[origin=c]{90}{FGE}}
          & suddenness & suddenness \\
          & familiarity & familiarity \\
          & expectedness & predict\_event \\
          & pleasantness & pleasantness \\
          & goal relevance & goal\_relevance \\
          & agent\_intention & chance\_responsibility \\
          & self\_cause & self\_responsibility \\
          & agent\_cause & other\_responsibility \\
          & certainty & predict\_consequence \\
          & goal\_consistency & goal\_support \\
          & control & self\_control \\
          & coping & accept\_consequence \\
          & self\_consistency & standards \\
          & moral & social\_norms \\
          & attention & attention \\
          \midrule
          \multirow{16}{*}{\rotatebox[origin=c]{90}{CovidET}}
          & familiarity & familiarity \\
          & expectedness & predict\_event \\
          & pleasantness & pleasantness \\
          & goal relevance & goal\_relevance \\
          & self-responsibility & self\_responsibility \\
          & other-responsibility & other\_responsibility \\
          & predictability & predict\_consequence \\
          & goal conduciveness & goal\_support \\
          & self-controllable & self\_control \\
          & other-controllable & other\_control \\
          & circumstances-controllable & chance\_control \\
          & problem-focused coping & accept\_consequence \\
          & consistency with internal values & standards \\
          & consistency with social norms & social\_norms \\
          & attentional activity & attention \\
          & effort & effort \\
          \bottomrule
      \end{tabular}
  }
  \caption{Mapping of Appraisal Dimensions. \textit{"Original"} indicates the appraisal dimensions in the original dataset, while \textit{"EnVent"} indicates the mapped appraisal dimensions in the EnVent dataset.}
  \label{app:tab:app_mapping}
\end{table}

\subsection{Unifying Appraisal Dimensions}
\label{app:app_mapping}
As the definition of appraisal dimensions vary across EnVent, FGE, and CovidET, we use the 21 appraisal dimensions from EnVent as anchor and manually mapped appraisal dimensions from FGE and CovidET to those of EnVent. The details of appraisal dimension mapping is shown in Table~\ref{app:tab:app_mapping}. 

Further, the appraisal ratings in FGE and CovidET follow a 10-point likert scale instead of the 5-point scale in EnVent. To unify the scale, we normalize the ratings to a 5-point scale by dividing the ratings by 2 and rounding to the nearest integer.

\section{Data Examples}
\label{app:data_examples}

We provide examples of data instances from the EnVent, FGE, and CovidET datasets, including the event description and appraisal ratings. We additionally include the demographic information provided in EnVent dataset, which we utilized to carry out experiments \S\ref{sec:results_and_analysis}.

\begin{tcolorbox}[
  enhanced,
  attach boxed title to top center={
      yshift=-3mm,yshifttext=-1mm
  },
  breakable,
  colframe=green2!75!black,
  colbacktitle=green2,
  title=Example from EnVent,
  coltitle=black,
  collower=blue, 
  label=tomi_examples,
  fonttitle=\bfseries,
  boxed title style={size=small,colframe=yellow!50!black}
  ]
  \textbf{<Event>} 

  People get under my skin. Like for example if an entitled customer shows up at my work and  demands to speak to my manager for a simple issue that I can resolve. This happens on almost a daily occurrence and it really makes me angry. \\

  \textbf{<Appraisal Ratings>} \\
  \aboverulesep=0ex
  \belowrulesep=0ex
  \begin{tabular}{ |c|c| }
    \toprule
    \textbf{Appraisal Dimension} & \textbf{Rating} \\
    \midrule
    suddenness & 2 \\
    \midrule
    familiarity & 5 \\
    \midrule
    predict\_event & 5 \\
    \midrule 
    ... & ... \\
    \bottomrule
  \end{tabular}
  \tcblower
  \textbf{<Demographic Information>} \\
  \resizebox{\columnwidth}{!}{
    \aboverulesep=0ex
    \belowrulesep=0ex
    \begin{tabular}{ |p{0.3\columnwidth}|p{0.7\columnwidth}| }
      \toprule
      \textbf{Info Type} & \textbf{Value} \\
      \midrule
      previous participation & Yes, first time, I will answer the following questions. \\
      \midrule
      age & 18 \\
      \midrule
      gender & Male \\
      \midrule
      education & High school \\
      \midrule
      ethnicity & North American \\
      \midrule
      extravert & 2.0 \\
      \midrule
      critical & 1.0 \\
      \midrule
      ... & ... \\
      \bottomrule
    \end{tabular}
  }
\end{tcolorbox}

\begin{tcolorbox}[
  enhanced,
  attach boxed title to top center={
      yshift=-3mm,yshifttext=-1mm
  },
  breakable,
  colframe=green2!75!black,
  colbacktitle=green2,
  title=Example from FGE,
  coltitle=black,
  collower=purple, 
  label=tomi_examples,
  fonttitle=\bfseries,
  boxed title style={size=small,colframe=yellow!50!black}
  ]
  \textbf{<Event>} \\
  NAMEVAR was very lost in her Organic Chemistry class, but when she looked at the other students, she seemed to be the only one struggling with the material. She had to raise her hand to ask the professor to explain the problem a second time. The other students in the class chuckled. \\

  \textbf{<Appraisal Ratings>} \\
  \aboverulesep=0ex
  \belowrulesep=0ex
  \begin{tabular}{ |c|c| }
    \toprule
    \textbf{Appraisal Dimension} & \textbf{Rating} \\
    \midrule
    predict\_event & 2 \\
    \midrule
    pleasantness & 1 \\
    \midrule
    goal\_support & 5 \\
    \midrule
    other\_responsibility & 1 \\
    \midrule
    ... & ... \\
    \bottomrule
  \end{tabular}

\end{tcolorbox}

\begin{tcolorbox}[
  enhanced,
  attach boxed title to top center={
      yshift=-3mm,yshifttext=-1mm
  },
  breakable,
  colframe=green2!75!black,
  colbacktitle=green2,
  title=Example from CovidET,
  coltitle=black,
  collower=purple, 
  label=tomi_examples,
  fonttitle=\bfseries,
  boxed title style={size=small,colframe=yellow!50!black}
  ]
  \textbf{<Event>} \\
  I don't even know how to speak of this grief. I have read of many stories of people losing their loved ones, but it didn't happen to my family until today. I lost my uncle who is my best friend, we lived abroad all our lives but because of the pandemic I returned home, and barely managed to see him those 2 years. He was the kindest and purest human being. I met him briefly yesterday as we took him to the hospital. Today he passed away from COVID, he was deathly afraid from needles and the vaccination. I feel so so powerless. So guilty I didn't reply to his phone call 3 days earlier as my family was also sick of Covid and I was caring for them. The death feels like it could have been avoided, he took the remaining precautions but it didn't work. At least he died happy alongside his family in his hometown where I live. I don't know what or how to live life without him in my mind, without meeting him ever again, with seeing the same places without him. Im tired, I want to cry. We both hated living at home now I am stuck living on earth without him, I feel this home is the right place for me - was waiting eagerly for him to return. But he barely lasted a week before his death. But mentally I am exhausted of living here, of the pandemic, of having my friends all abroad and getting out of contact, of being alone to face his death. To face this life. Life feels so tasteless. \\

  \textbf{<Appraisal Rating>} \\
  \aboverulesep=0ex
  \belowrulesep=0ex
  \begin{tabular}{ |c|c| }
    \toprule
    \textbf{Appraisal Dimension} & \textbf{Rating} \\
    \midrule
    self\_responsibility & 2 \\
    \midrule
    other\_responsibility & 2 \\
    \midrule
    accept\_consequence & 1 \\
    \midrule
    goal\_relevance & 5 \\
    \midrule
    attention & 2 \\
    \midrule
    ... & ... \\
    \bottomrule
  \end{tabular}
\end{tcolorbox}

\section{Annotation Validity}
\label{app:quality_control}

Each benchmark includes quality-control procedures to ensure annotation validity.
We summarize below the quality assurance protocols reported in the original papers for each datasets.

\textbf{EnVent}~\citep{hofmann2020appraisal}: Data quality was ensured through multiple steps. Participants were restricted to native English speakers from select countries (US, UK, Australia, New Zealand, Canada, or Ireland) with a Prolific approval rate of $\geq80\%$. Two types of attention checks were included: one requiring the selection of a specific scale point, and another involving a manual text response. Surveys were restricted to desktop devices to avoid mobile-based auto-corrections. The data collection process was structured in nine rounds, including an initial pilot to gather feedback and adjust instructions, and later rounds to balance under-represented data. Additional care was taken for dimensions which difficulty for annotators to reliably differentiate.

\textbf{FGE}~\citep{skerry2015neural}: Participants rated how involved the named character was in the narrative. Participants with an average score below 7 were excluded, and individual responses with scores below 5 were discarded. This led to the exclusion of 22 participants, resulting in a final pool of 238 valid subjects and an average of 15.4 responses per stimulus across 200 items. This attention check served as a proxy for engagement and comprehension of the story content.

\textbf{CovidET}~\citep{zhan2023evaluating}: Includes detailed inter-rater agreement metrics: Krippendorff’s alpha (interval) yielded 0.647, indicating substantial agreement; average Spearman’s $\rho$ across dimensions was 0.497 (statistically significant); and the mean absolute difference between annotators was low (1.734 on a 1–9 scale). These results indicate that even in a subjective task, annotators exhibited consistent patterns in their judgments.

It is worth noting that CovidET includes limited number of annotators per sample.
We incorporate this dataset for two main reasons.
First, to evaluate model generalizability on out-of-domain scenario: whereas EnVent and FGE consist of daliy event descriptions, CovidET specifically focuses on situations related to COVID-19, providing a different context for testing the robustness of language models.
Second, to examine how the number of annotations per instance affects model performance. 
We observed that model predictions exhibited similar trends in datasets with a moderate number of annotators, whereas the CovidET dataset showed different model behavior.

\section{Analysis of Human Appraisal Ratings}
\label{app:human_rating_analysis}
To determine the shape of the distribution for CADE-LSM, we conduct analysis on the mode of distribution of appraisal ratings. As shown in Table~\ref{app:tab:modality}, regardless of the appraisal dimension, the majority of the appraisal ratings follow some unimodal distribution. In addition, some appraisal ratings display a bimodal distribution. As such, we take two approaches in CADE-LSM where we model the appraisal ratings as a unimodal distribution or as a mixture of two unimodal distributions, effectively reproducing the bimodal distribution.

\begin{table*}
  \centering
  \resizebox{0.75\textwidth}{!}{
    \begin{tabular}{cccccc}
    \toprule
    Dimension & \textit{Unimodal} & \textit{Bimodal} & \textit{Trimodal} & \textit{Quadmodal} & \textit{Pentamodal} \\
    \midrule
    suddenness & 716 & 468 & 0 & 0 & 16 \\
    familiarity & 701 & 484 & 0 & 0 & 15 \\
    predict\_event & 695 & 493 & 0 & 0 & 12 \\
    pleasantness & 905 & 293 & 0 & 0 & 2 \\
    unpleasantness & 856 & 339 & 0 & 0 & 5 \\
    goal\_relevance & 725 & 462 & 0 & 0 & 13 \\
    chance\_responsibility & 813 & 372 & 0 & 0 & 15 \\
    self\_responsibility & 798 & 394 & 0 & 0 & 8 \\
    other\_responsibility & 793 & 396 & 0 & 0 & 11 \\
    predict\_consequence & 686 & 497 & 0 & 0 & 17 \\
    goal\_support & 786 & 398 & 0 & 0 & 16 \\
    urgency & 734 & 446 & 0 & 0 & 20 \\
    self\_control & 700 & 489 & 0 & 0 & 11 \\
    other\_control & 734 & 453 & 0 & 0 & 13 \\
    chance\_control & 784 & 405 & 0 & 0 & 11 \\
    accept\_consequence & 675 & 506 & 0 & 0 & 19 \\
    standards & 843 & 345 & 0 & 0 & 12 \\
    social\_norms & 941 & 251 & 0 & 0 & 8 \\
    attention & 662 & 513 & 0 & 0 & 25 \\
    not\_consider & 819 & 374 & 0 & 0 & 7 \\
    effort & 712 & 476 & 0 & 0 & 12 \\
    \bottomrule
    \end{tabular}
  }
  \caption{Modality of distribution of appraisal ratings in the EnVent dataset. As each appraisal dimension of each situation is rated by five different annotators, we calculate the modality of the distribution ranging from 1 (unimodal) to 5 (uniform).}
  \label{app:tab:modality}
\end{table*}

\section{Prompts}
\label{app:prompts}
Here we list the prompt templates we used to conduct prompting experiments with LLMs. We use the \textit{vanilla} prompt for experiments without auxiliary (demographic) information:

\begin{tcolorbox}[
    enhanced,
    attach boxed title to top center={
        yshift=-3mm,yshifttext=-1mm
    },
    breakable,
    colframe=green1!75!black,
    colbacktitle=green1,
    title=Vanilla Prompt Template,
    coltitle=black,
    collower=black, 
    label=tomi_examples,
    fonttitle=\bfseries,
    boxed title style={size=small,colframe=green1!50!black}
    ]
    \textbf{<System Prompt>} \\
    Put yourself in the shoes of the writer at the time when the event happened, and try to reconstruct how that [Situation] was perceived. How much do these statements apply? (1 means “Not at all” and 5 means “Extremely”) \\
    \textbf{</System Prompt>}
    \tcblower
    \textbf{<User Prompt>} \\
    Put yourself in the shoes of the writer at the time when the event happened, and try to reconstruct how that [Situation] was perceived. How much do these statements apply? Please rate the situation according to the statements using the Likert scale. The scale ranges from 1 to 5 where 1 means 'Not at all' and 5 means '“Extremely”'. Provide your rating in the following format: "Rating: [Score]". Do not add any explanation or elaboration to your answer. \\
    
    [Situation] \\ 
    \{\{context\}\} \\
    
    [Experiencer's Feeling] \\
    \{\{statements\}\} \\
    \textbf{</User Prompt>}
\end{tcolorbox}

When conducting experiments that involve auxiliary information such as demographic information and personality traits, we first flatten the structured information into natural language description. Specifically, for demographic information (as shown in the EnVent data example in \S~\ref{app:data_examples}), we use the demographic information to fill the following template:
\begin{align*}
&\texttt{You are a \{age\} years old} \\
&\texttt{\{ethnicity\} \{gender\} whose} \\
&\texttt{education level is "\{education\}".}
\end{align*}
For instance, if a given demographic information is 
\begin{align*}
    &\texttt{Age = 28} \\
    &\texttt{Ethnicity = African} \\
    &\texttt{Gender = Female} \\
    &\texttt{Education = College}
\end{align*}
it will be converted into "
\texttt{You are a 28 years old African female whose education level is "college".}"
We incorporate demographic information into the context by filling the demographic description into the \texttt{\{\{demographic info\}\}} slot in the following prompt:

\begin{tcolorbox}[
    enhanced,
    attach boxed title to top center={
        yshift=-3mm,yshifttext=-1mm
    },
    breakable,
    colframe=green1!75!black,
    colbacktitle=green1,
    title=Prompt Template (+Demo),
    coltitle=black,
    collower=black, 
    label=tomi_examples,
    fonttitle=\bfseries,
    boxed title style={size=small,colframe=green1!50!black}
    ]
    \textbf{<System Prompt>} \\
    \{\{demographic info\}\} Put yourself in the shoes of the writer at the time when the event happened, and try to reconstruct how that [Situation] was perceived. How much do these statements apply? (1 means “Not at all” and 5 means “Extremely”) \\
    \textbf{</System Prompt>}
    \tcblower
    \textbf{<User Prompt>} \\
     \{\{demographic info\}\} Put yourself in the shoes of the writer at the time when the event happened, and try to reconstruct how that [Situation] was perceived. How much do these statements apply? Please rate the situation according to the statements using the Likert scale. The scale ranges from 1 to 5 where 1 means 'Not at all' and 5 means '“Extremely”'. Provide your rating in the following format: "Rating: [Score]". Do not add any explanation or elaboration to your answer.\\
     
     [Situation] \\
     \{\{context\}\} \\
     
     [Experiencer's Feeling] \\
     \{\{statements\}\} \\
    \textbf{</User Prompt>}
\end{tcolorbox}

Similar to demographic information, we convert the structured personality traits information into natural language descriptions. In the EnVent dataset, the authors used the Big-Five personality traits \citep{costa1999five}. Specifically, the authors leveraged the 10-item assessment of the Big-Five personality traits \citep{gosling2003very}. For each trait of the Big-Five, two descriptions are rated to reflect the annotator's tendency:

\allowdisplaybreaks
\begin{align*}
    \texttt{Openness to experience}& \\
    &\texttt{open} \\
    &\texttt{conventional} \\ 
    \\
    \texttt{Conscientiousness}& \\
    &\texttt{dependable} \\
    &\texttt{disorganized} \\
    \\
    \texttt{Extraversion}& \\
    &\texttt{extraverted} \\
    &\texttt{quiet} \\
    \\
    \texttt{Agreeableness} & \\
    &\texttt{sympathetic} \\
    &\texttt{critical} \\
    \\
    \texttt{Emotional Stability} & \\
    &\texttt{calm} \\
    &\texttt{anxious}
\end{align*}

Each of the 10 items are rated on a scale of 1-7. We compare the rating of the paired items (e.g. comparing \texttt{open} and \texttt{conventional} for \texttt{openness to experience}) and select the item with a higher rating as the description. We omit the personality trait if there is a tie in the rating of the corresponding items. We use the selected item description to fill the following template:
\begin{align*}
    &\texttt{You are a \{openness\}} \\
    &\texttt{\{conscientiousness\}\{extraversion\}\}} \\
    &\texttt{\{agreeableness\}\{emotional\_stability\}} \\
    &\texttt{person.}
\end{align*}

We incorporate personality trait information by replacing the \texttt{\{\{personality traits\}\}} placeholder with the filled personality trait template:

\begin{tcolorbox}[
    enhanced,
    attach boxed title to top center={
        yshift=-3mm,yshifttext=-1mm
    },
    breakable,
    colframe=green1!75!black,
    colbacktitle=green1,
    title=Prompt Template (+Traits),
    coltitle=black,
    collower=black, 
    label=tomi_examples,
    fonttitle=\bfseries,
    boxed title style={size=small,colframe=green1!50!black}
    ]
    \textbf{<System Prompt>} \\
    \{\{personality traits\}\} Put yourself in the shoes of the writer at the time when the event happened, and try to reconstruct how that [Situation] was perceived. How much do these statements apply? (1 means “Not at all” and 5 means “Extremely”) \\
    \textbf{</System Prompt>}
    \tcblower
    \textbf{<User Prompt>} \\
     \{\{personality traits\}\} Put yourself in the shoes of the writer at the time when the event happened, and try to reconstruct how that [Situation] was perceived. How much do these statements apply? Please rate the situation according to the statements using the Likert scale. The scale ranges from 1 to 5 where 1 means 'Not at all' and 5 means '“Extremely”'. Provide your rating in the following format: "Rating: [Score]". Do not add any explanation or elaboration to your answer.\\
     
     [Situation] \\
     \{\{context\}\}\\
     
     [Experiencer's Feeling]\\
     \{\{statements\}\} \\
    \textbf{</User Prompt>}
\end{tcolorbox}

We use the following prompt to incorporate both the demographic information and personality trait information:

\begin{tcolorbox}[
    enhanced,
    attach boxed title to top center={
        yshift=-3mm,yshifttext=-1mm
    },
    breakable,
    colframe=green1!75!black,
    colbacktitle=green1,
    title=Prompt Template (+Demo +Traits),
    coltitle=black,
    collower=black, 
    label=tomi_examples,
    fonttitle=\bfseries,
    boxed title style={size=small,colframe=green1!50!black}
    ]
    \textbf{<System Prompt>} \\
    \{\{demographic info\}\} \{\{personality traits\}\} Put yourself in the shoes of the writer at the time when the event happened, and try to reconstruct how that [Situation] was perceived. How much do these statements apply? (1 means “Not at all” and 5 means “Extremely”) \\
    \textbf{</System Prompt>}
    \tcblower
    \textbf{<User Prompt>} \\
    \{\{demographic info\}\} \{\{personality traits\}\} Put yourself in the shoes of the writer at the time when the event happened, and try to reconstruct how that [Situation] was perceived. How much do these statements apply? Please rate the situation according to the statements using the Likert scale. The scale ranges from 1 to 5 where 1 means 'Not at all' and 5 means '“Extremely”'. Provide your rating in the following format: "Rating: [Score]". Do not add any explanation or elaboration to your answer. \\
    
    [Situation] \\
    \{\{context\}\} \\
    
    [Experiencer's Feeling] \\
    \{\{statements\}\} \\
    \textbf{</User Prompt>}
\end{tcolorbox}

\section{Unimodal and Bimodal Label Smoothing}
\label{app:unimodal_vs_bimodal}
As discussed in \S\ref{sec:method}, in addition to smoothing the appraisal ratings assuming an unimodal shape, we also conducted experiments by assuming that the distribution over ratings follows a bimodal distribution.
To simulate a bimodal distribution, we conduct label smoothing using a mixture of two discretized Gaussian distributions, one centered at the ground truth rating and the other at a non-adjacent rating\footnote{For instance, if the ground truth rating is "\texttt{2}", we would allocate the majority of probability density to ratings "\texttt{2}" and one of ["\texttt{4}", "\texttt{5}"].}.
Experiment results demonstrate that the bimodal CADE-LSM model underperforms that with an unimodal assumption in Wasserstein distance and $\mu$-MAE and achieved minor improvement in $\sigma^2$-MAE. We present the comparison in Table~\ref{app:tab:unimodal_vs_bimodal}

\begin{table} [h!]
  \centering 
  \resizebox{\columnwidth}{!}{
      \begin{tabular}{c c c c}
          \toprule
          Model & Wasserstein$\downarrow$ & $\mu$-MAE$\downarrow$ & $\sigma^2$-MAE$\downarrow$ \\ 
          \midrule 
          CADE-LSM-\texttt{Unimodal} & 0.773 & 0.665 & 0.837 \\
          CADE-LSM-\texttt{bimodal} & 0.900 & 0.793 &  0.782 \\
          \bottomrule
      \end{tabular}
  }
  \caption{Comparison of unimodal versus bimodal label smoothing results.}
  \label{app:tab:unimodal_vs_bimodal}
  \vspace{-1em}
\end{table}

\section{Analysis of LLMs Appraisal Ratings}
A key factor that influences the variance of LLMs' appraisal rating is the \texttt{temperature} parameter, which controls the kurtosis of the logits over the vocabulary space. To ensure a fair comparison among LLMs and fine-tuned autoencoding models, we conduct a grid search over temperatures ranging from [0, 1.5]. We provide the Wasserstein distance, $\mu$-MAE, and $\sigma^2$-MAE metrics with respect to the sampling temperature used in Table~\ref{app:tab:temperature_study}.
\begin{table} [t!]
  \centering 
  \resizebox{\columnwidth}{!}{
      \begin{tabular}{P{0.01\columnwidth} c c c c}
          \toprule
          & Temperature & Wasserstein$\downarrow$ & $\mu$-MAE$\downarrow$ & $\sigma^2$-MAE$\downarrow$ \\
          \midrule
          \multirow{6}{*}{\rotatebox[origin=c]{90}{Llama3.1-8B}}
          & 0.00 & 1.130 & 0.932 & 0.887 \\
          & \textbf{0.25} & \textbf{1.094} & \textbf{0.904} & 0.826 \\
          & 0.50 & 1.113 & 0.941 & 0.812 \\
          & 0.75 & 1.117 & 0.973 & 0.808 \\
          & 1.00 & 1.127 & 1.083 & 0.798 \\
          & 1.25 & 1.142 & 1.129 & \textbf{0.791} \\
          \midrule
          \multirow{6}{*}{\rotatebox[origin=c]{90}{Qwen2.5-7B}}
          & 0.00 & 1.144 & 0.984 & 0.888 \\
          & 0.25 & 1.117 & 0.960 & 0.861 \\
          & 0.50 & 1.094 & 0.945 & 0.826 \\
          & \textbf{0.75} & \textbf{1.078} & \textbf{0.919} & 0.817 \\
          & 1.00 & 1.084 & 0.939 & 0.814 \\
          & 1.25 & 1.090 & 0.950 & \textbf{0.811} \\
          \midrule
          \multirow{7}{*}{\rotatebox[origin=c]{90}{Llama3.3-70B}}
          & 0.00 & 1.070 & 0.956 & 0.882 \\
          & 0.25 & 1.061 & 0.958 & 0.873 \\
          & 0.50 & 1.053 & 0.958 & 0.862 \\
          & 0.75 & 1.042 & 0.951 & 0.851 \\
          & 1.00 & 1.032 & 0.942 & 0.840 \\
          & 1.25 & 1.022 & 0.935 & 0.830 \\
          & \textbf{1.50} & \textbf{1.012} & \textbf{0.926} & \textbf{0.820} \\
          \midrule
          \multirow{7}{*}{\rotatebox[origin=c]{90}{Qwen2.5-72B}}
          & 0.00 & 1.092 & 1.021 & 0.870 \\
          & 0.25 & 1.056 & 1.027 & 0.830 \\
          & 0.50 & 1.022 & 1.025 & 0.799 \\
          & 0.75 & 0.995 & 0.945 & 0.771 \\
          & 1.00 & 0.974 & 0.892 & 0.749 \\
          & 1.25 & 0.954 & 0.873 & 0.738 \\
          & \textbf{1.50} & \textbf{0.945} & \textbf{0.852} & \textbf{0.736} \\
          \bottomrule
      \end{tabular}
  }
  \caption{Temperature study results.}
  \label{app:tab:temperature_study}
  \vspace{-1em}
\end{table}

\begin{table*}[ht!]
  \centering
       \resizebox{0.9\linewidth}{!}{
  \setlength{\tabcolsep}{4mm}{
    \begin{tabular}{lcccccc}
    \toprule
    \multicolumn{1}{c}{\multirow{2}[4]{*}{\textbf{Dimension}}} & \multicolumn{3}{c}{\textbf{Llama3.1-8B}} & \multicolumn{3}{c}{\textbf{Qwen2.5-7B}} \\
\cmidrule{2-7}          & ~w.~Demo & ~w.~Traits & ~w.~Demo \textbackslash{}\& Traits  & ~w.~Demo & ~w.~Traits & ~w.~Demo \textbackslash{}\& Traits  \\
    \midrule
    \textit{suddenness} & 0.747 & \textbf{0.051} & 0.243 & 0.506 & 0.480 & 0.440 \\
    \textit{familiarity} & \textbf{0.000} & \textbf{0.000} & \textbf{0.000} & \textbf{0.000} & \textbf{0.000} & 0.700 \\
    \textit{predict\_event} & \textbf{0.001} & 0.236 & \textbf{0.008} & 0.515 & \textbf{0.086} & \textbf{0.067} \\
    \textit{pleasantness} & 0.372 & 0.885 & 0.936 & 0.616 & 0.351 & 0.825 \\
    \textit{unpleasantness} & 0.936 & 0.948 & 0.276 & 0.598 & 0.597 & 0.633 \\
    \textit{goal\_relevance} & 0.258 & \textbf{0.000} & 0.973 & \textbf{0.093} & 0.621 & \textbf{0.008} \\
    \textit{chance\_responsblt} & 0.419 & 0.756 & 0.187 & \textbf{0.070} & \textbf{0.034} & 0.424 \\
    \textit{self\_responsblt} & \textbf{0.047} & 0.100 & \textbf{0.020} & 0.273 & 0.135 & 0.122 \\
    \textit{other\_responsblt} & 0.183 & 0.632 & 0.792 & 0.452 & 0.308 & \textbf{0.046} \\
    \textit{predict\_conseq} & \textbf{0.010} & \textbf{0.000} & \textbf{0.034} & 0.909 & 0.173 & 0.754 \\
    \textit{self\_control} & \textbf{0.001} & \textbf{0.000} & \textbf{0.014} & 0.114 & \textbf{0.000} & \textbf{0.014} \\
    \textit{urgency} & 0.752 & \textbf{0.015} & 0.862 & 0.273 & \textbf{0.016} & \textbf{0.034} \\
    \textit{other\_control} & \textbf{0.000} & 0.435 & 0.105 & 0.965 & 0.273 & \textbf{0.019} \\
    \textit{chance\_control} & 0.115 & 0.101 & 0.069 & \textbf{0.100} & \textbf{0.008} & \textbf{0.001} \\
    \textit{accept\_conseq} & \textbf{0.015} & \textbf{0.022} & \textbf{0.042} & 0.264 & \textbf{0.015} & \textbf{0.000} \\
    \textit{standards} & 0.866 & \textbf{0.000} & 0.772 & 0.668 & \textbf{0.062} & 0.104 \\
    \textit{social\_norms} & 0.631 & 0.847 & 0.979 & 0.146 & \textbf{0.030} & \textbf{0.000} \\
    \textit{attention} & \textbf{0.033} & \textbf{0.005} & 0.755 & 0.702 & 0.498 & 0.357 \\
    \textit{not\_consider} & 0.272 & \textbf{0.000} & 0.952 & 0.637 & 0.752 & 0.269 \\
    \textit{effort} & 0.576 & \textbf{0.001} & \textbf{0.017} & 0.641 & 0.610 & \textbf{0.000} \\
    \textit{goal\_support} & 0.983 & 0.548 & 0.530 & 0.151 & 0.421 & 0.516 \\
    \bottomrule
    \end{tabular}}}%
      \caption{Statistical significance of personal profile incorporation}
  \label{tab:sig_profile}%
\end{table*}%

\section{Implementation details}
\paragraph{CADE-VAE:}
\label{sec:apdx_cadevae}
Given the contextualized representation $\bm{h}=\text{PLM}(\bm{s})$ for each situation $\bm{s}$.
We compute the approximation variational posterior $q_{\phi}(\bm{z}|\bm{h})$ using the inference network $\Phi (\bm{h}; \phi)$:
\begin{equation}
 \begin{aligned}
  &\bm{\mu} = \bm{W}_{\mu}\bm{h}+\bm{b}_{\mu}\\ 
 	&\log \bm{\sigma}^2=\bm{W}_\sigma\bm{h}+\bm{b}_{\sigma}\\
 	&\bm{z}=\bm{\mu}+\bm{\sigma} \odot \bm{\epsilon}
 \end{aligned}
 \label{equ:vae}
\end{equation}
where $\bm{W}_{\mu}$, $\bm{W}_{\sigma}$, $\bm{b}_{\mu}$, and $\bm{b}_{\sigma}$ are parameters for two MLPs.
$\bm{\mu}$ and $\bm{\sigma}$ define a multivariate Gaussian distribution with a diagonal covariance matrix, and $\bm{\epsilon} \sim \mathcal{N}(0,\textbf{I} )$.
Then, we sample from $q_{\phi}(\bm{z}|\bm{h}) \simeq \mathcal{N}(\bm{\mu},\,\bm{\sigma}^{2}\textbf{I})$ to generate 
$\bm{z} \in \mathbb R^l$ as the latent representation, where $l$ is the dimension of the representation.
We use inference network $\Phi(\bm{h}; \phi)$ for inferring $\bm{z}$ and two-layer parameterized MLP $\Theta(\bm{h};\theta)$ as the decoder to reconstruct $\bm{h}$.
The parameters can be optimized by maximizing the evidence lower bound (ELBO):

\begin{equation}
 \mathcal L_{vae} = \mathbb{E}_{\Phi}[\log \Theta(\bm{h};\theta)]-\text{KL}(\Phi||p(\bm{z}))
\end{equation}
where~$p(\bm{z})$ is the prior follows the Gaussian distribution $\mathcal{N}(0,\textbf{I})$.
We also introduce a regression loss, $\mathcal L_{\text{regression}} = \text{MSE}(y, \bm{h})$ to predict appraisal scores, enhancing text representation for more efficient distribution estimation. 
The overall objective function is a multi-task learning objective:
\begin{equation}
	\mathcal L =  \mathcal L_{vae} +\lambda \mathcal L_{\text{regression}}
\end{equation} 
where $\lambda$ is the coefficient that balances the contribution of each component in the training process.

\section{Statistical Significance of Personal Profile Incorporation}
\label{sec:apdx_sg}
We have conducted a one-tailed two-sample T-test to analyze the influence of personal profiles to the distribution of ratings produced by LLMs. 
As shown in Table~\ref{tab:sig_profile}, we found that 1/3 of the appraisal dimensions exhibit statistically significant improvements when either personality traits or demographical information are integrated, across both models. 
Specifically, adding personality traits improves performance on \textit{familiarity, self\_control, urgency, accept\_consequence}, and \textit{standards}.
Incorporating demographical information yields significant gains on \textit{familiarity}.
When both types of personal profiles are integrated, improvements are observed for \textit{predict\_event, self\_control, accept\_consequence}, and \textit{effort}.

For dimensions that significantly improved after adding personality traits, many have strong correlation with certain traits. For instance, \textit{self\_control}, which is defined as ``the experiencer expects positive consequences for themselves", is highly relevant to traits such as ``anxious" and ``calm".
For dimensions that significantly improved after adding demographic information, the ``age", ``gender", ``education", and ``ethnicity" information aid the model to possess a more concrete portrait of the user, hence easier to determine \textit{familiarity}, which represents whether ``the event was familiar to its experiencer". LLMs do generally acquire, via pre-training, a strong prior knowledge as to whether a person from a particular demographic group would be familiar with a given situation.
For instance, the event ``England scored in the 2nd minute of the Euros final" would be more familiar for people of ``European" ethnicity compared to people of ``East Asian" ethnicity.

\section{Qualitative Study}
\label{sec:apdx_qs}
\begin{table*}[h!]
  \centering
     \resizebox{0.9\linewidth}{!}{
  \setlength{\tabcolsep}{4mm}{
    \begin{tabular}{lccccc}
    \toprule
    \multicolumn{1}{c}{\multirow{2}[4]{*}{\textbf{Models}}} & \multicolumn{5}{c}{Top Quantified Appraisal Dimensions} \\
\cmidrule{2-6}          & Rank 1 & Rank 2 & Rank 3 & Rank 4 & Rank 5 \\
    \midrule
    CADE-LS & \textit{pleasantness} & \textit{unpleasantness} & \textit{social\_norms} & \textit{standards} & \textit{self\_responsibility} \\
    ~~~~~~~~~~w. Demo & \textit{pleasantness} & \textit{social\_norms} & \textit{unpleasantness} & \textit{standards} & \textit{self\_responsibility} \\
    ~~~~~~~~~~w. Traits & \textit{pleasantness} & \textit{unpleasantness} & \textit{social\_norms} & \textit{self\_responsibility} & \textit{standards} \\
    ~~~~~~~~~~w. Demo \& Traits  & \textit{pleasantness} & \textit{unpleasantness} & \textit{social\_norms} & \textit{self\_responsibility} & \textit{standards} \\
    \midrule
    \midrule
    Llama3-8B & \textit{social\_norms} & \textit{pleasantness} & \textit{unpleasantness} & \textit{self\_control} & \textit{chance\_responsibility} \\
    ~~~~~~~~~~w. Demo & \textit{social\_norms} & \textit{pleasantness} & \textit{unpleasantness} & \textit{self\_control} & \textit{Attention} \\
    ~~~~~~~~~~w. Traits & \textit{social\_norms} & \textit{pleasantness} & \textit{unpleasantness} & \textit{self\_control} & \textit{Attention} \\
    ~~~~~~~~~~w. Demo \& Traits  & \textit{social\_norms} & \textit{pleasantness} & \textit{unpleasantness} & \textit{self\_control} & \textit{Attention} \\
    \midrule
    \midrule
    Qwen2.5-7B & \textit{pleasantness} & \textit{unpleasantness} & \textit{social\_norms} & \textit{self\_control} & \textit{Attention} \\
    ~~~~~~~~~~w. Demo & \textit{pleasantness} & \textit{social\_norms} & \textit{unpleasantness} & \textit{self\_control} & \textit{Attention} \\
    ~~~~~~~~~~w. Traits & \textit{pleasantness} & \textit{social\_norms} & \textit{unpleasantness} & \textit{self\_control} & \textit{Attention} \\
    ~~~~~~~~~~w. Demo \& Traits  & \textit{pleasantness} & \textit{social\_norms} & \textit{unpleasantness} & \textit{self\_control} & \textit{Attention} \\
    \bottomrule
    \end{tabular}}}%
    \caption{Qualitative analysis of subjectivity in well-modeled appraisal dimensions across various models.}
  \label{tab:quali_ad_top}%
\end{table*}%

\begin{table*}[h!]
  \centering
     \resizebox{0.9\linewidth}{!}{
  \setlength{\tabcolsep}{4mm}{
    \begin{tabular}{lccccc}
    \toprule
    \multicolumn{1}{c}{\multirow{2}[4]{*}{\textbf{Models}}} & \multicolumn{5}{c}{Bottom Quantified Appraisal Dimensions} \\
\cmidrule{2-6}          & Rank 1 & Rank 2 & Rank 3 & Rank 4 & Rank 5 \\
    \midrule
    CADE-LS & \textit{urgency} & \textit{accept\_consequence} & \textit{predict\_consequence} & \textit{goal\_support} & \textit{self\_control} \\
    ~~~~~~~~~~w. Demo & \textit{urgency} & \textit{goal\_support} & \textit{accept\_consequence} & \textit{predict\_consequence} & \textit{self\_control} \\
    ~~~~~~~~~~w. Traits & \textit{goal\_support} & \textit{urgency} & \textit{predict\_consequence} & \textit{self\_control} & \textit{accept\_consequence} \\
    ~~~~~~~~~~w. Demo \& Traits  & \textit{predict\_event} & \textit{urgency} & \textit{predict\_consequence} & \textit{accept\_consequence} & \textit{self\_control} \\
    \midrule
    \midrule
    Llama3-8B & \textit{other\_control} & \textit{goal\_support} & \textit{familiarity} & \textit{predict\_consequence} & \textit{chance\_control} \\
    ~~~~~~~~~~w. Demo & \textit{other\_control} & \textit{goal\_support} & \textit{familiarity} & \textit{predict\_consequence} & \textit{chance\_control} \\
    ~~~~~~~~~~w. Traits & \textit{other\_control} & \textit{goal\_support} & \textit{familiarity} & \textit{predict\_consequence} & \textit{chance\_control} \\
    ~~~~~~~~~~w. Demo \& Traits  & \textit{other\_control} & \textit{goal\_support} & \textit{familiarity} & \textit{predict\_consequence} & \textit{chance\_control} \\
    \midrule
    \midrule
    Qwen2.5-7B & \textit{familiarity} & \textit{goal\_support} & \textit{other\_control} & \textit{predict\_event} & \textit{accept\_consequence} \\
    ~~~~~~~~~~w. Demo & \textit{familiarity} & \textit{goal\_support} & \textit{other\_control} & \textit{predict\_event} & \textit{chance\_control} \\
    ~~~~~~~~~~w. Traits & \textit{familiarity} & \textit{goal\_support} & \textit{other\_control} & \textit{predict\_event} & \textit{chance\_control} \\
    ~~~~~~~~~~w. Demo \& Traits  & \textit{familiarity} & \textit{goal\_support} & \textit{other\_control} & \textit{predict\_event} & \textit{chance\_control} \\
    \bottomrule
    \end{tabular}}}%
    \caption{Qualitative analysis of subjectivity in poorly modeled appraisal dimensions across various models.}
  \label{tab:quali_ad_btm}%
\end{table*}%

 \begin{figure*}[ht!]
 	\centering
 	{\includegraphics[width=0.8\textwidth]{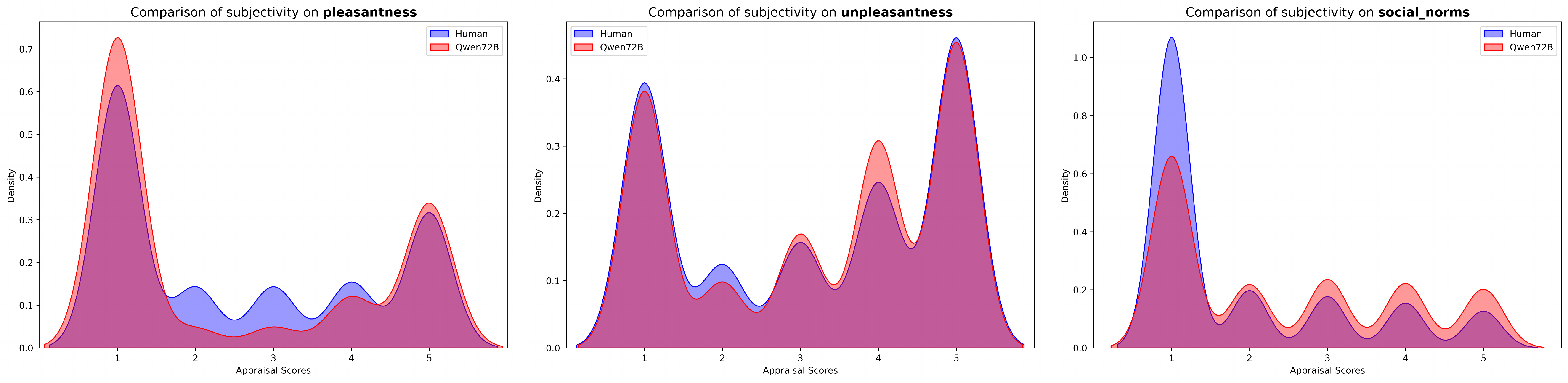}}
 	\caption{Comparison of appraisal distributions for top-ranked subjectivity dimensions between human annotators and Qwen-72B.}
  \label{fig:top_subj}
 \end{figure*}

 \begin{figure*}[ht!]
 	\centering
 	{\includegraphics[width=0.8\textwidth]{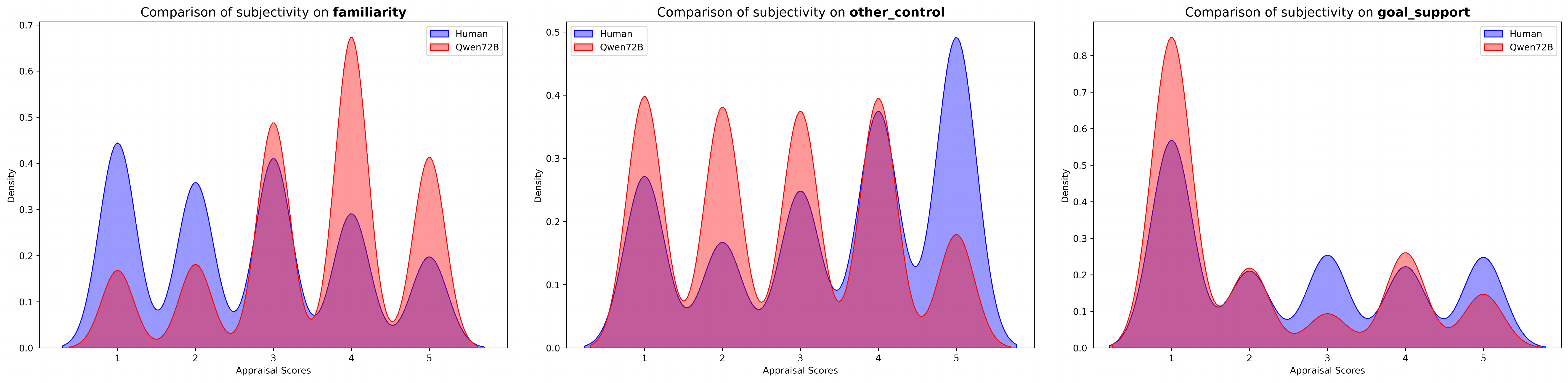}}
 	\caption{Comparison of appraisal distributions for bottom-ranked subjectivity dimensions between human annotators and Qwen-72B.}
  \label{fig:btm_subj}
 \end{figure*}
\paragraph{Understanding subjectivity from various appraisal dimensions.}
Table~\ref{tab:quali_ad_top} and \ref{tab:quali_ad_btm} present the qualitative study on subjectivity across well-modeled and poorly modeled appraisal dimensions across various models.
Figure~\ref{fig:top_subj} and \ref{fig:btm_subj} show the appraisal distributions predicted by LLMs compared to human annotators for the selected top-ranked and bottom-ranked subjectivity dimensions.

\begin{figure}[t!]
	\centering
	\includegraphics[width=\columnwidth]{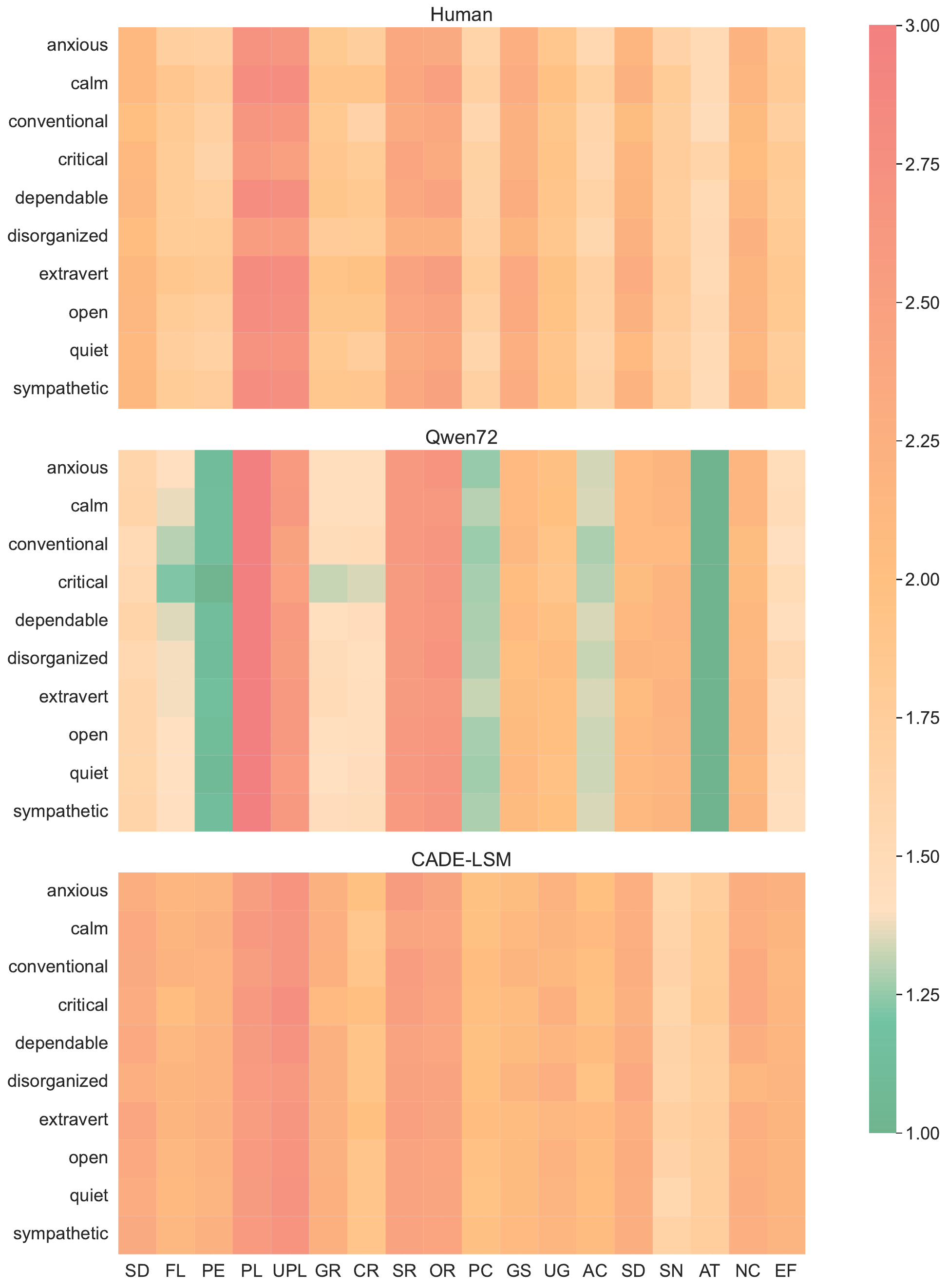}
	\caption{Comparison of appraisal variance in different personalities in the EnVent dataset.}
	\label{fig:trait_var}
	\vspace{-1em}
\end{figure}

\paragraph{Effects of Personality Traits}
We conduct a qualitative analysis to investigate the effects of various personality traits on the cognitive subjectivity.
As shown in Figure~\ref{fig:trait_var}, certain appraisal dimensions such as \textit{pleasantness} and \textit{unpleasantness} show inherent variance regardless of individuals’ personality traits.
These variance can largely be attributed to the characteristics of situation descriptions used in EnVent~\citep{hofmann2020appraisal}.
Specifically, the annotation manual indicates that repeated appraisal ratings are conducted on experiencer-reported situations, which often carry polarized sentiments, as people tend to recall events that evoke strong emotional reactions.
I addition, annotators were instructed to describe an event that made them feel one out of twelve predefined emotions, which were deliberately designed to span both postive and negative sentiments.
As a results, the reported situations naturally contain bi-polar sentiment, futher amplifying variance in \textit{pleasantness} and \textit{unpleasantness}.
%

\section{Correlations between mean and variance}
\label{app:mean_var_corr}
\begin{table}[t!]
    \centering
    \resizebox{\linewidth}{!}{
    \begin{tabular}{ c c c c }
    \toprule
    Model & Metric1 ($m1$) & Metric2 ($m2$) & $\rho_{m1, m2}$ \\ 
    \toprule
    Llama3.3-8B	& $\sigma^2$-MAE & $\mu$-MAE & 0.596 \\
    Llama3.3-8B	& $\sigma^2$-MAE & Wasserstein & 0.683 \\
    \midrule 
    Qwen2.5-7B & $\sigma^2$-MAE & $\mu$-MAE & 0.292 \\
    Qwen2.5-7B	& $\sigma^2$-MAE & Wasserstein & 0.538 \\
    \midrule 
    Llama3.1-70B & $\sigma^2$-MAE & $\mu$-MAE & 0.832 \\
    Llama3.1-70B & $\sigma^2$-MAE & Wasserstein & 0.922 \\
    \midrule 
    Qwen2.5-72B	& $\sigma^2$-MAE & $\mu$-MAE & 0.369 \\
    Qwen2.5-72B	& $\sigma^2$-MAE & Wasserstein & 0.544 \\
    \bottomrule
    \end{tabular}
    }
    \caption{Pearson correlation between the two point-estimate metrics across 4 models. }
    \label{tab:metric_corr}
    \vspace{-1em}
\end{table}

\paragraph{\textit{Correlations between mean and variance is model-dependent}} 
To examine whether proficiency in modeling average rating tendency implies competency in capturing subjectivity, we conduct a correlation analysis between the two metrics across various models. 
We evaluate models using two point-estimate metrics: $\mu$-MAE, which measures a model’s ability to capture the population’s average rating, and $\sigma^2$-MAE, which quantifies its ability to capture rating subjectivity within the population. 
Results from Table~\ref{tab:metric_corr} show that there exists a weak correlation between the two metrics except for the Llama3.1-70B model. Therefore, in addition to making models for representative of the population (e.g. $\mu$-MAE), it is equally important to make them more capable of modeling the nuanced subjectivity within each population (e.g. $\sigma^2$-MAE).

\end{document}